\documentclass[preprint,12pt]{elsarticle}

\usepackage[letterpaper,top=2cm,bottom=2cm,left=3cm,right=3cm,marginparwidth=1.75cm]{geometry}

\usepackage{amsmath}
\usepackage{graphicx}
\usepackage{float}
\usepackage{amssymb}
\usepackage{xcolor}
\usepackage{tikz}
\usepackage{subcaption}
\usepackage{titlesec}
\titleformat{\paragraph}[block]{\normalfont\normalsize\bfseries}{\theparagraph}{1em}{}

\usetikzlibrary{arrows.meta,positioning}

\journal{Nuclear Physics B}
\begin{document}

\begin{frontmatter}



\title{Application of Reduced-Order Models for Temporal Multiscale Representations in the Prediction of Dynamical Systems}


\author[inst1]{Elias Al Ghazal\corref{cor1}}
\ead{elias.al_ghazal@ensam.eu}
\author[inst1]{Jad Mounayer}
\author[inst1]{Beatriz Moya}
\author[inst1]{Sebastian Rodriguez}
\author[inst2]{Chady Ghnatios}
\author[inst1,inst3]{Francisco Chinesta}

\cortext[cor1]{Corresponding author}

\affiliation[inst1]{organization={PIMM Lab, ENSAM Institute of Technology},%
            city={Paris},
            postcode={75013},
            country={France}}

\affiliation[inst2]{organization={Department of Mechanical Engineering, University of North Florida},%
            addressline={1 UNF Drive}, 
            city={Jacksonville},
            state={FL},
            postcode={32224}, 
            country={United States of America}}

\affiliation[inst3]{organization={CNRS@CREATE LTD},%
            city={Singapore},
            country={Singapore}}

\begin{abstract}
Modeling and predicting the dynamics of complex multiscale systems remains a significant challenge due to their inherent nonlinearities and sensitivity to initial conditions, as well as limitations of traditional machine learning methods that fail to capture high frequency behaviours. To overcome these difficulties, we propose three approaches for multiscale learning. The first leverages the Partition of Unity (PU) method, integrated with neural networks, to decompose the dynamics into local components and directly predict both macro- and micro-scale behaviors. The second applies  the Singular Value Decomposition (SVD) to extract dominant modes that explicitly separate macro- and micro-scale dynamics. Since full access to the data matrix is rarely available in practice, we further employ a Sparse High-Order SVD to reconstruct multiscale dynamics from limited measurements. Together, these approaches ensure that both coarse and fine dynamics are accurately captured, making the framework effective for real-world applications involving complex, multi-scale phenomena and adaptable to higher-dimensional systems with incomplete observations, by providing an approximation and interpretation in all time scales present in the phenomena under study.
\end{abstract}

\begin{graphicalabstract}
\begin{center}
\vspace*{1.25cm}
\begin{tikzpicture}[node distance=2cm, auto, thick, >=latex, font=\small]

\definecolor{pu}{RGB}{66,135,245}
\definecolor{svd}{RGB}{245,166,35}
\definecolor{hosvd}{RGB}{120,180,75}

\node[draw, rounded corners=8pt, fill=gray!15, minimum width=3cm, minimum height=1cm, align=center] (input) {Multiscale \\ system data};

\node[draw, rounded corners=8pt, fill=pu!30, minimum width=3cm, minimum height=1cm, align=center, below left=1.5cm and 1 of input] (pu) {Partition of Unity \\ Neural Framework};
\node[draw, rounded corners=8pt, fill=svd!30, minimum width=3cm, minimum height=1cm, align=center, below=1.5cm of input] (svd) {Singular Value \\ Decomposition};
\node[draw, rounded corners=8pt, fill=hosvd!30, minimum width=3cm, minimum height=1cm, align=center, below right=1.5cm and 1cm of input] (hosvd) {Sparse High Order \\ SVD Reconstruction};

\node[draw, rounded corners=8pt, fill=gray!15, minimum width=3cm, minimum height=1cm, align=center, below=1.5cm of svd] (output) {Reconstructed / Predicted \\ Multiscale Dynamics};

\draw[->] (input) -- (pu);
\draw[->] (input) -- (svd);
\draw[->] (input) -- (hosvd);
\draw[->] (pu) -- (output);
\draw[->] (svd) -- (output);
\draw[->] (hosvd) -- (output);

\end{tikzpicture}
\end{center}
\end{graphicalabstract}

\begin{highlights}


\item Development of a multi-scale learning framework---accounting for the presence of slow and fast dynamics---for dynamical systems, using the Partition of Unity and Reduced Order Models to capture the full system behavior and mitigate the over-smoothing tendency of machine learning approximations.
\item Reconstruction of functions governing the evolution of dynamical systems, ranging from cases with full access to the state space to scenarios with limited and sparse measurements, using techniques applicable to high-dimensional spaces.

\end{highlights}

\begin{keyword}



Multiscale dynamics \sep Partition of Unity \sep Neural networks \sep Singular Value Decomposition \sep Sparse High Order SVD \sep Model reduction \sep Data-driven modeling

\end{keyword}

\end{frontmatter}
\newpage

\section{Introduction}

Understanding the dynamics of systems governed by nonlinear behaviours is essential in many scientific and engineering fields, as it enables the prediction and control of complex processes \cite{wang2017data}. Particularly, whether forecasting the weather, designing mechanical systems, or analyzing financial markets, accurately modeling and predicting systems characterized by multiple time scales remains a significant challenge. These systems exhibit a rich interplay between slow, long-term trends and fast, short-lived events, a phenomenon known as multi-scale dynamics \cite{ibanez2019multiscale}. Capturing both fast and slow dynamics simultaneously is further complicated by the inherent nonlinearity of many systems \cite{li2023learning,michel2019multiple}. For example, in climate modeling, long-term trends such as global warming (coarse dynamics) interact with short-term fluctuations like weather changes (fine dynamics) \cite{bordoni2025futures}. These time scales are deeply intertwined, with short-term events influencing long-term trends and vice versa, making it difficult to capture the full scope of the system's behavior in a single model or approximation. The complexity and interplay of fast and slow processes can lead to chaotic patterns that are difficult to predict or understand without sophisticated techniques. Similarly, in mechanical systems, the interaction between high-frequency oscillations and low-frequency movements requires careful consideration of both fine and coarse dynamics. Such systems can exhibit sudden, dramatic changes in behavior depending on initial conditions or minor perturbations \cite{pilipchuk1998sensitive}. These challenges are amplified when the governing function $\dot{x} = f(x)$ that describes the system's dynamics is unknown.



When $\dot{x} = f(x)$ is known, traditional techniques such as numerical integration can be employed to solve the resulting differential equations and model the time evolution of the system’s state $x(t)$. However, in many practical applications, the exact form of $f(x)$ is unknown, and only discrete observations of the system’s state are available at different time points \cite{Raissi2019}. In such cases, the challenge shifts to inferring, or learning, the unknown forcing function $f(x)$ that drives the system’s evolution. In machine learning, this is particularly challenging because the system is often observed as time-series data, and the goal is to discover the underlying dynamics that govern its behavior.

Several data-driven black-box machine learning methods have been proposed for this purpose \cite{Chen2022,ZHANG2004283}. However, model discovery is unstable, specially in multiscale scenarios, and can lead to explosion or vanishing of gradients \cite{CHAKRABORTY2024117442}. To regularize the process, some methods leverage sparsity-promoting techniques to identify parsimonious models that accurately capture system dynamics, even in the presence of noise or limited data. However, these models may struggle when system dynamics involve multiple time scales or when the Fourier spectrum of $f(x)$ contains both low and high frequencies \cite{zheng2024les}. As stated in \cite{karniadakis2021physics}, physics-informed machine learning increases stability effects by leading the learning of dynamical system approximations towards local minima closer to the absolute minimum solution due to the imposition of physical restrictions, such as PDEs \cite{Raissi2019}, thermodynamics \cite{urdeitx2025comparison,tierz2025feasibility,hernandez2021structure}, and other types of biases \cite{ yu2024learning}. Moreover, structures such as Koopman operators \cite{baikonode,nathaniel2025deep,otto2021koopman, Ghnatios2024LearningTransformed}, Sparse Identification of Nonlinear Dynamics (SINDy) \cite{brunton2016discovering}, and Neural Operators \cite{park2024dynamical,li2025d,michalowska2024neural}, with DeepONets \cite{cong2025transfer,lin2021operator,goswami2022physics} being a particularly appealing option, have demonstrated great efficiency in this task. Nonetheless, these methods are susceptible to overfitting to the dominant low-frequency scales, thereby neglecting the high-frequency phenomena. 

This is due to the fact that, in spite of the use of learning and physics biases, neural networks learn inherently low frequencies, and learning the fast frequencies tends to be part of a overfitting issue due to the optimization process. Authors in \cite{rahaman2019spectral} showcase how deep neaural networks suffers from a strong bias to low frequencies, which are learned faster than the high frequencies. These results are also supported by Zhi-Qin John Xu et al. \cite{xu2019frequency}.

Most solutions to this challenge propose learning in the frequency domain \cite{xu2019frequency}, where Fourier neural
operators (FNO) is one of the main methods in this field \cite{cong2025transfer,wen2022u}. However, it is not always possible to learn efficiently in the frequency domain. In dynamical systems governed by $\dot{x} = f(x)$ the evolution of the state may depend solely on its current value through the function $f(x)$, rather than explicitly on time. Consequently, learning this problem in the frequency domain is challenging since the behaviour is defined in the state space and not in the temporal domain. 

Following this idea, and as a solution for the oversmoothing issues of machine learning, recent advances in dynamical systems modeling have focused on separating the dynamics into distinct levels: coarse (long-term) and fine (short-term) scales \cite{Rodriguez2025,Pasquale2021,RODRIGUEZ2024107277}, proposing a sort of hierarchical learning scheme\cite{brenner2024learning}. By isolating these scales, each can be modeled independently, improving both accuracy and interpretability \cite{RODRIGUEZ2024107277}. This separation allows for a more precise understanding of how different time scales interact, without the computational inefficiencies and inaccuracies that arise from modeling them together \cite{ibanez2019multiscale}. 

The Multi-scale Deep Neural Network (MscaleDNN), for instance, 
captures features at multiple spatial scales \cite{liu2020multi}. In the field of physics-informed machine learning, we also find works dedicated to the treatment of different scales \cite{wang2024multi,liu2021multiscale,ahmed2023multifidelity}. Such is the case of Wu et al. \cite{wu2024capturing}, which solves time-dependent linear transport equations using the so-called Asymptotic-Preserving Convolutional Deep Operator Networks (APCONs). In this case, separation of scales is performed by using the Knudsen number. However, in many cases, the identification of scales is not apparent or may remain undetermined in the absence of effective separation strategies. Often, only a bifurcation between two scales is performed, despite the fact that each scale may comprise multiple principal modes.

One effective technique for this separation is the use of structure-preserving and partitioning strategies that decompose the system’s dynamics into local approximations valid within specific regions of the domain \cite{Beddig2023}, such as in the case of the Partition of Unity (PU) \cite{ibanez2019multiscale}. By leveraging these approaches, both coarse and fine dynamics can be represented separately, ensuring that the model focuses on the relevant scales at each level and improving modeling accuracy and efficiency.

Also, reduced-order models aim to address this issue from a multiscaling approach by emphasizing the emergence of the principal modes inherent in the problems by contributing to the refinement of the learning process by unveiling the dynamic patterns inherent within its modes \cite{baker2023learning,sentz2025reduced}.

In this paper, we showcase the use of three different approaches to model the macro-and micro-scale dynamics of complex systems, improving both the accuracy and efficiency of predictions. These approaches are motivated by two key challenges: (i) the inherently multiscale nature of the problems, and (ii) their highly nonlinear dynamics with strong sensitivity to initial conditions. The first method employed is the Partition of Unity (PU), which is traditionally used for function decomposition in order to study local effects. In this case, the decomposition itself is learned from data to learn and reconstruct the original $f(x)$ by capturing both local variability and global structure \cite{Rodriguez2025}. The second approach tested in this work is the SVD-based approach to extract both macro- and micro-scale dynamics through dominant modes. The third approach proposed allows us to utilize a Sparse High-Order SVD, which allows learning multiscale modes from sparse measurements and provides a foundation for extending these techniques to higher-dimensional systems. Collectively, these contributions improve the accuracy and efficiency in the learning process of $f(x)$ without a priori knowledge of the separation of scales for complex nonlinear systems with strong sensitivity to initial conditions.

Consequently, this work is structured as follows. Section 2 elaborates on the use of PU for the approximation of multiscale dynamical systems. Section 3 presents the use of Singular Value Decomposition (SVD) for multiscale function approximations. Finally, In Section 4 we introduce the use of a spare-high order SVD learned with neural networks hierarchically to approximate the different modes and scales of which the function is composed of, leading to a method to tackle also multiscale learning in multidimensional problems from sparse measurements.

\section{Partition of Unity Method}

\begin{figure}[h!]
    \centering
    
    \begin{subfigure}{0.95\textwidth}
        \centering
        \begin{tikzpicture}[>=Stealth, scale=0.9, every node/.style={transform shape}]
            \draw[-] (0,0) -- (8,0);

            \draw[very thick, purple] (0,1.5) -- (2,0);
            \node[black] at (-0.5,1.7) {$N_1(x)$};

            \draw[very thick, orange] (0,0) -- (2,1.5) -- (4,0);
            \node[black] at (1.2,1.7) {$N_2(x)$};

            \draw[very thick, red] (2,0) -- (4,1.5) -- (6,0);
            \node[black] at (3.2,1.7) {$N_3(x)$};

            \draw[very thick, green!70!black] (4,0) -- (6,1.5) -- (8,0);
            \node[black] at (5.2,1.7) {$N_4(x)$};

            \draw[very thick, brown] (6,0) -- (8,1.5);
            \node[black] at (8.5,1.7) {$N_5(x)$};

            \foreach \x/\label in {0/x_1, 2/x_2, 4/x_3, 6/x_4, 8/x_5} {
                \filldraw[blue] (\x,0) circle (2.5pt) node[below, black, yshift=-2pt] {$\label$};
            };
            
            \begin{scope}[yshift=2cm] 
            \draw[thick, blue, domain=0:8, samples=100, smooth] 
                plot (\x, {0.5*sin(360*\x/6) + 1});
        
            \foreach \x/\n in {0/1, 4/3, 8/5} {
                \coordinate (P) at (\x,{0.5*sin(360*\x/6 ) + 1});
                \fill[blue] (P) circle (2.5pt);
                \node[above] at (P) {$F_{\n}$};
            }
        \end{scope}

            \node[blue] at (10.4,2.1) {\textbf{MACRO SCALE}};

            \begin{scope}[yshift=-3.5cm]  
            \draw[thick] (0,0) -- (4,0);
            \node[below] at (0,-0.2) {$x_1$};
            \node[below] at (2,-0.2) {$x_2$};
            \node[below] at (4,-0.2) {$x_3$};

            \foreach \x in {0,0.5,1,1.5,2,2.5,3,3.5,4} {
                \filldraw[orange] (\x,0) circle (2pt);
            }

        \begin{scope}[yshift=0.5cm] 
            \draw[thick, orange, domain=0:4, samples=100, smooth] 
            plot (\x, {0.3*sin(360*\x-2)});
            \node[orange] at (4.65,0.1) {$E_2(x)$};
        \end{scope}

    \end{scope}

    \draw[->, thick, dashed] (0,-0.5) -- (0,-3.4);
    \draw[->, thick, dashed] (4,-0.5) -- (4,-3.4);

            \begin{scope}[yshift=-1.8cm]
                \draw[thick] (2,0) -- (6,0);
                \node[below] at (2,-0.2) {$x_2$};
                \node[below] at (4,-0.2) {$x_3$};
                \node[below] at (6,-0.2) {$x_4$};

                \foreach \x in {2.0,2.4,2.8,3.2,3.6,4.0,4.4,4.8,5.2,5.6,6.0} {
                    \filldraw[red] (\x,0) circle (2pt);
                };
                
                \begin{scope}[yshift=-0.10cm] 
                \draw[thick, red, domain=2:6, samples=200, smooth] 
                plot (\x, {0.3*sin(360*((\x-2))) + 0.8});
                \node[red] at (6.65,0.8) {$E_3(x)$};
            \end{scope}

            \end{scope}

            \draw[->, thick, dashed] (2,-0.5) -- (2,-1.6);
            \draw[->, thick, dashed] (6,-0.5) -- (6,-1.6);
            
        \node[orange!50!red] at (10.4,-2) {\textbf{MICRO SCALE}};
        \end{tikzpicture}
        \caption{Shape functions of a multiscale approach. (Top) Macro shape functions provide coarse-scale approximation. (Bottom) Micro shape functions capture fine-scale details.}
        \label{fig:macro_micro_shape}
    \end{subfigure}
    
    \vspace{1cm}
    
    \begin{subfigure}{0.95\textwidth}
        \centering
        \begin{tikzpicture}[>=Stealth, every node/.style={align=center}]
            \node[draw, thick, rounded corners, fill=gray!15, minimum width=3cm, minimum height=1cm] (fx) at (0,5) {$f(x)$};

            \node[draw, thick, rounded corners, fill=blue!15, minimum width=3cm, minimum height=1cm] (macro) at (-3,2.5) {Macro-scale \\ \scriptsize ANN predicts $F_i^m$};

            \node[draw, thick, rounded corners, fill=red!15, minimum width=3cm, minimum height=1cm] (micro) at (3,2.5) {Micro-scale \\ \scriptsize Trainable vector predicts $E_i^m(x)$};

            \node[draw, thick, rounded corners, fill=black!10, minimum width=6cm, minimum height=1cm] (out) at (0,0) {Predicted $f(x) \approx$ \eqref{eq:multi_enrichment} \\ \scriptsize $\sum_{m,i} F_i^m N_i(x) E_i^m(x)$};

            \draw[->, thick] (fx.south) -- (macro.north);
            \draw[->, thick] (fx.south) -- (micro.north);

            \node[coordinate] (join) at (0,1.2) {};
            \draw[-, thick] (macro.south) -- ++(0,-0.8) -- (join);
            \draw[-, thick] (micro.south) -- ++(0,-0.8) -- (join);

            \draw[->, thick] (join) -- (out.north);
        \end{tikzpicture}
        \caption{Decomposition of $f(x)$ into macro- and micro-scale components. Macro-scale dynamics are predicted by a neural network (ANN), while micro-scale dynamics are captured by a trainable vector. 
        The two components are combined to reconstruct the enriched approximation of $f(x)$.}
        \label{fig:pu_workflow}
    \end{subfigure}
    
    \caption{Workflow for Partition of Unity.}
    \label{fig:multiscale_combined}
\end{figure}
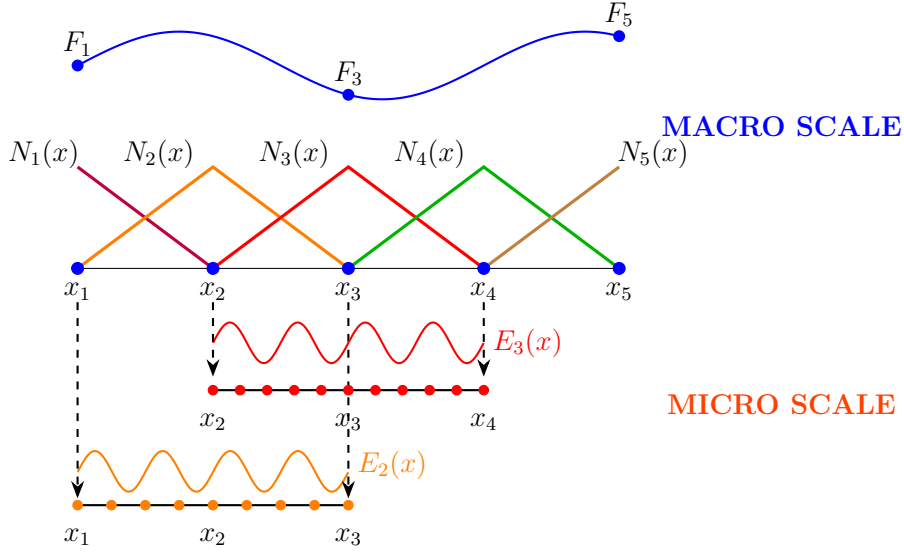
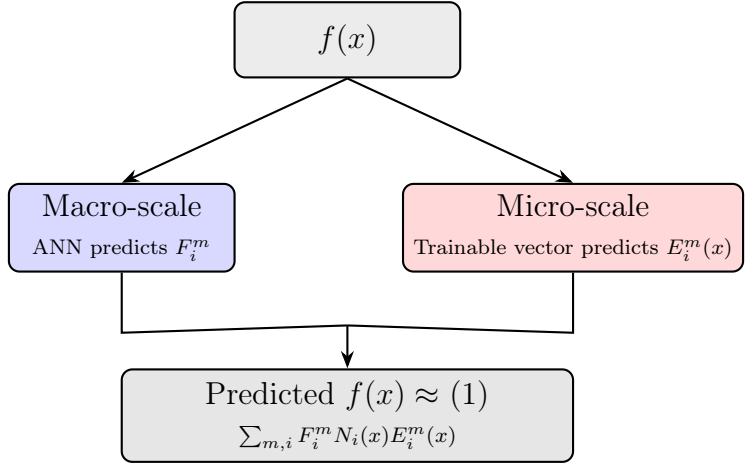

The Partition of Unity (PU) method \cite{MELENK1996289}, widely used in computational mechanics, offers an effective means of handling multiscale dynamics in dynamical systems. Traditional models often struggle to capture the complex interactions between scales, and this limitation is further exacerbated in many machine learning approaches due to the well-known spectral bias issue \cite{rahaman2019spectralbiasneuralnetworks}. The problem is especially pronounced when the system exhibits nonlinear and high-frequency behavior.

To address these challenges, we propose a framework that integrates the PU method with machine learning. Specifically, we use a neural network to model the macro-scale dynamics and a trainable vector to capture fine-scale enrichments (as shown in Fig.~\ref{fig:multiscale_combined}).

\subsection{Coarse (Macro) Approximation}

We begin with a coarse approximation of the governing dynamics $\dot{x} = f(x)$, represented as:

\[
f(x) = \sum_{i=1}^{N} F_i N_i(x),
\]

where $N_i(x)$ are piecewise linear shape functions that are equal to 1 at the macro node $x_i$ and 0 at the neighboring macro nodes. Here, $F_i$ is the nodal macro value corresponding to the macro node $x_i$ (see the top of Figure~\ref{fig:macro_micro_shape}). This formulation provides a coarse-scale, macro-level representation of the system dynamics. However, it has limited ability to capture fine-scale transients.

\subsection{Fine-Scale (Micro) Enrichment}
To address this limitation, the coarse approximation is augmented with a fine-scale enrichment that enhances the resolution of the system’s dynamics at selected locations \cite{ibanez2019multiscale}. The enriched approximation can be written as:
\begin{align*}
f(x) &= \sum_{i=1}^{N} F_i\, N_i(x)\, E_i(x), \\
E_i(x) &=
\begin{cases}
G(x - x_i), & \text{if } x \in \Omega_i, \\
0, & \text{otherwise}.
\end{cases}
\end{align*}

and \( \Omega_i \) is a local micro-domain centered at macro node \( x_i \), defined to be twice the size of the macro element (see the bottom of Figure~\ref{fig:macro_micro_shape}). 

Although using a single enrichment function \( G(x) \) is computationally efficient, it may not suffice to resolve the richness of the dynamics in complex systems. To overcome this limitation, a parsimonious multi-enrichment approach can be adopted. This approach is inspired by methods such as the Proper Generalized Decomposition (PGD) \cite{GHNATIOS201229}, which allow for the decomposition of the system’s dynamics into multiple modes. The multi-enrichment formulation is given by:
\begin{equation}
f(x) = \sum_{m=1}^{M} \sum_{i=1}^{N} F_i^m\, N_i(x)\, G^m(x - x_i)
\label{eq:multi_enrichment}
\end{equation}

In this formulation, \( M \) represents the number of enrichment functions, and \( F^{m}_{i} \) and \( G^m(x - x_i) \) are the unknown coefficients and enrichment functions, respectively, for each enrichment level \( m \). The multi-enrichment approach captures the different frequencies present in the system’s dynamics by considering multiple levels of enrichment, with each level corresponding to a different frequency band. The first mode (\( m = 1 \)) captures the low-frequency dynamics, while subsequent modes capture higher-frequency components. This hierarchical enrichment process allows the model to efficiently resolve multi-scale dynamics in a manner that progressively captures more detailed features of the system's behavior.

\subsection{Learning Procedure}

To effectively learn the dynamics of the system from data, we parameterize the enrichment-based formulation using machine learning models. Specifically, we decompose the representation into two learnable components (see Figure~\ref{fig:pu_workflow}):

\begin{itemize}
  \item \textbf{Macro-scale component:} The macro coefficients \( F_i^m \) for each enrichment level \( m \) are modeled as outputs of a neural network with the corresponding \( x_i \). The network captures the coarse-scale, low-frequency behavior of the system and learns a smooth, global representation across the domain.
  
  \item \textbf{Micro-scale component:} Each enrichment function \( E_i(x) \) is defined over a small micro-domain centered at \( x_i \). The function depends only on the local coordinates relative to the macro element, rather than the absolute input \( x \), which allows the same micro-scale representation to be repeated across all macro elements. Accordingly, \( E_i^m \) is learned as a shared trainable vector, with gradients computed independently of any inputs.
 
\end{itemize}

The key idea is that the macro component provides a smooth global approximation, while the micro component enhances local resolution using a compact, shared vector. This separation ensures that high-resolution details do not overwhelm the neural network's capacity, allowing it to generalize better across the domain.

During training, both the neural network parameters and the enrichment vectors are jointly optimized by minimizing a loss function that compares the predicted output with the actual data.

This learning approach leverages the structure-preserving properties of the Partition of Unity method while allowing for efficient, data-driven learning of multiscale system dynamics. The result is a compact and interpretable model that captures both global trends and localized transients with high fidelity.

\subsection{Numerical results}

In this section, we present some numerical results obtained by applying the Partition of Unity (PU) method to model and predict the dynamics of several types of dynamical systems. For each of these systems, data is generated by simulating their evolution under various initial conditions. The conditions span a broad range of states, ensuring that the model is exposed to diverse trajectories and system dynamics. Each data point consists of pairs \( (x, \dot{x}) \), where \( x \) represents the state of the system and \( \dot{x} \) represents its time derivative.

The dataset is split into a training set (80\% of the data) and a testing set (20\% of the data). The training set is used to train the model, while the testing set is used to evaluate the model’s predictive performance. Prior to splitting, the data is shuffled randomly to avoid any bias due to the ordering of the data.

We tested the Partition of Unity method using different numbers of modes, including setups with 1, 2, or more macro and micro modes. These configurations allow us to assess the impact of including different modes on the model’s ability to capture both the global and fine-scale dynamics. By using a combination of both macro and micro modes, the model is able to capture large-scale global trends as well as fine-scale local fluctuations.

To determine the appropriate number of modes in the Partition of Unity framework, we adopt an adaptive enrichment strategy. Specifically, the model is trained incrementally by adding one mode at a time. After each enrichment step, the model’s performance is evaluated using the mean squared error (MSE) computed on the validation set. If the MSE remains above a predefined threshold of \(10^{-2}\), an additional mode is introduced to further enhance the model’s representational capacity. This process continues until the MSE falls below the threshold, indicating that the model has achieved a satisfactory level of accuracy. In this way, the enrichment process is guided by a quantitative stopping criterion rather than a fixed number of modes, ensuring an optimal balance between model accuracy and computational efficiency.

In our implementation, the macro components of the PU model are represented using feedforward neural networks. The macro neural network consists of three fully connected layers: the first maps the input to 64 hidden units, followed by a ReLU activation; the second maps to 32 hidden units with another ReLU; and the final layer outputs a scalar value. The micro components are modeled as trainable vectors that are directly optimized during training.

Training is performed over 300 epochs using the Adam optimizer with a learning rate of \(10^{-3}\) and a weight decay of \(10^{-4}\). The loss function used to train the model is a normalized relative error defined as:
\[
\text{Loss}(\mathbf{x}, \mathbf{y}) = \frac{ \| \mathbf{x} - \mathbf{y} \|_2 }{ \| \mathbf{y} \|_2 } \times 100,
\]
where \( \mathbf{x} \) is the model output and \( \mathbf{y} \) is the ground truth, and \( \| \cdot \|_2 \) denotes the \(L^2\) norm.

After training, the model is first evaluated on the held-out testing set. To further assess its generalization ability, we then test the model using only the initial condition $x_0$. From this starting point, the trained model predicts the system’s evolution over time via Runge–Kutta integration. The resulting trajectories are compared with the corresponding ground-truth dynamics.

We begin with a simple example and gradually move to more complex systems. The aim is to evaluate the performance of the method in capturing both the macro and micro scales of the system's behavior while maintaining the computational efficiency.

\subsubsection{A first example}\label{firstexample}
\begin{figure}[h!]
    \centering

    \begin{subfigure}{0.7\textwidth}
        \centering
        \includegraphics[width=\textwidth]{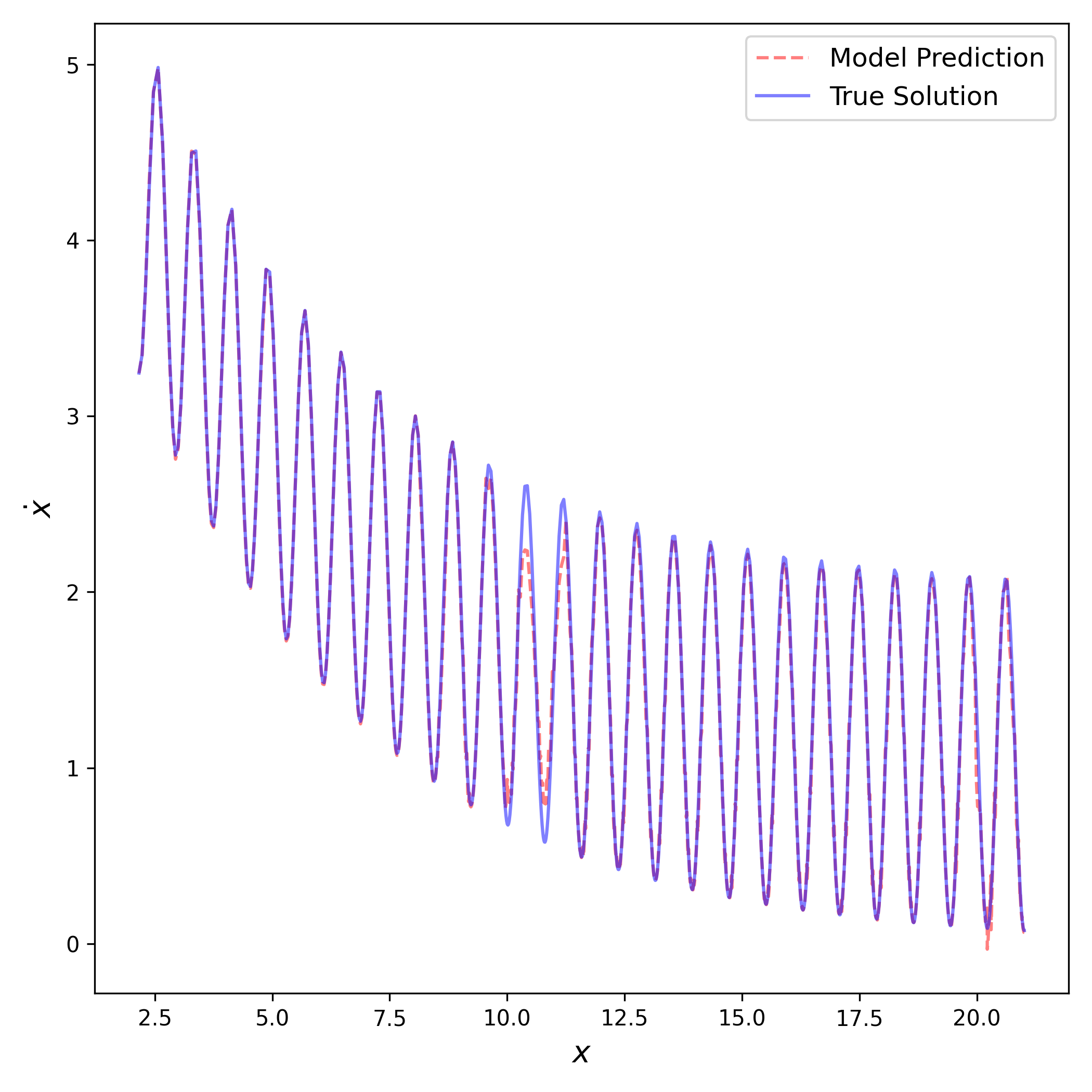}
        \caption{Function plot for Eq.~\eqref{eq:sin&exp}.}
        \label{fig:function1}
    \end{subfigure}
    
    \vspace{0.5cm}

    \makebox[\textwidth][c]{
        \begin{subfigure}{0.48\textwidth}
            \centering
            \includegraphics[width=\textwidth]{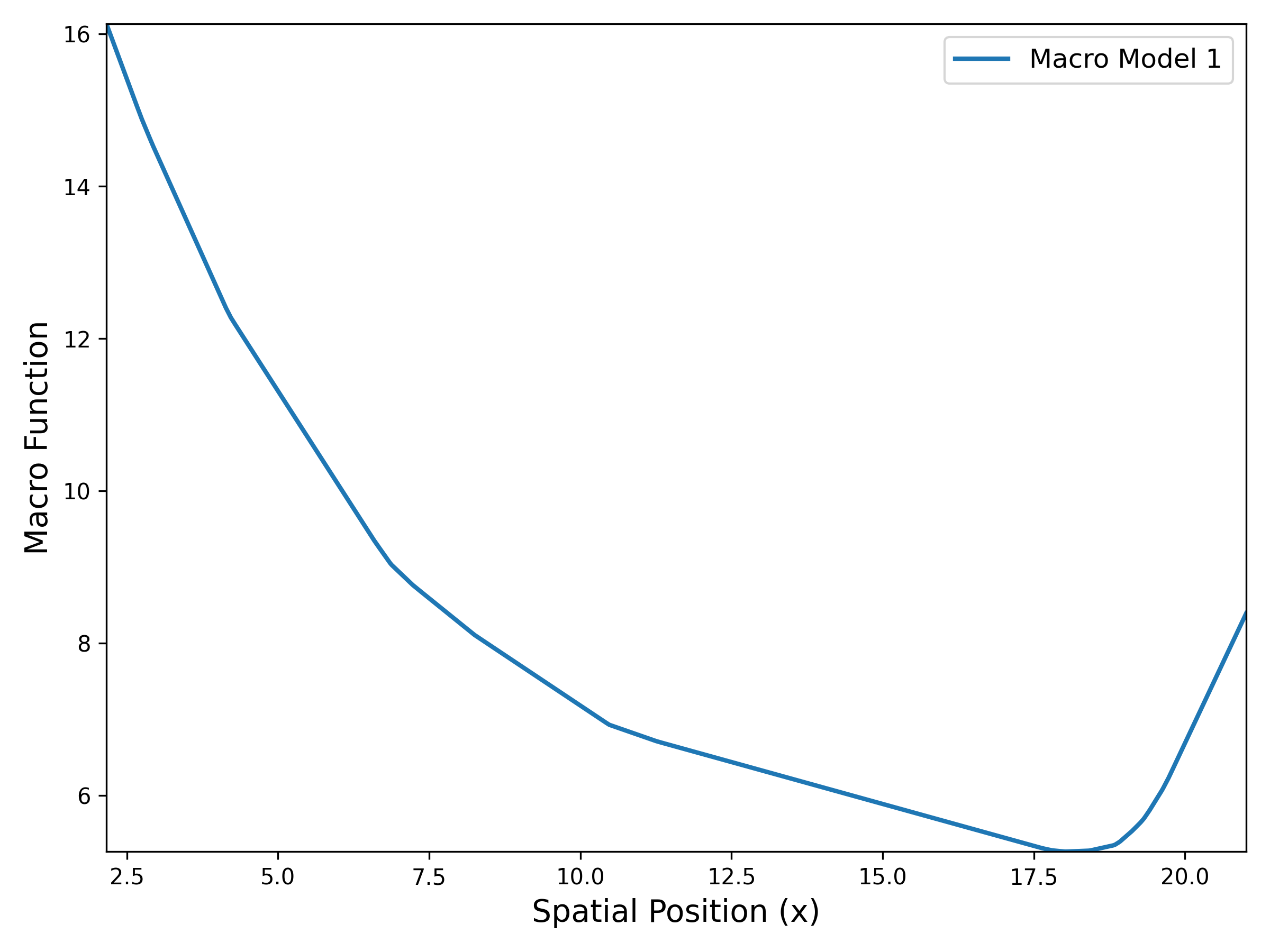}
            \caption{Macro component of Eq.~\eqref{eq:sin&exp}.}
            \label{fig:macro1}
        \end{subfigure}
        \hspace{0.04\textwidth}
        \begin{subfigure}{0.48\textwidth}
            \centering
            \includegraphics[width=\textwidth]{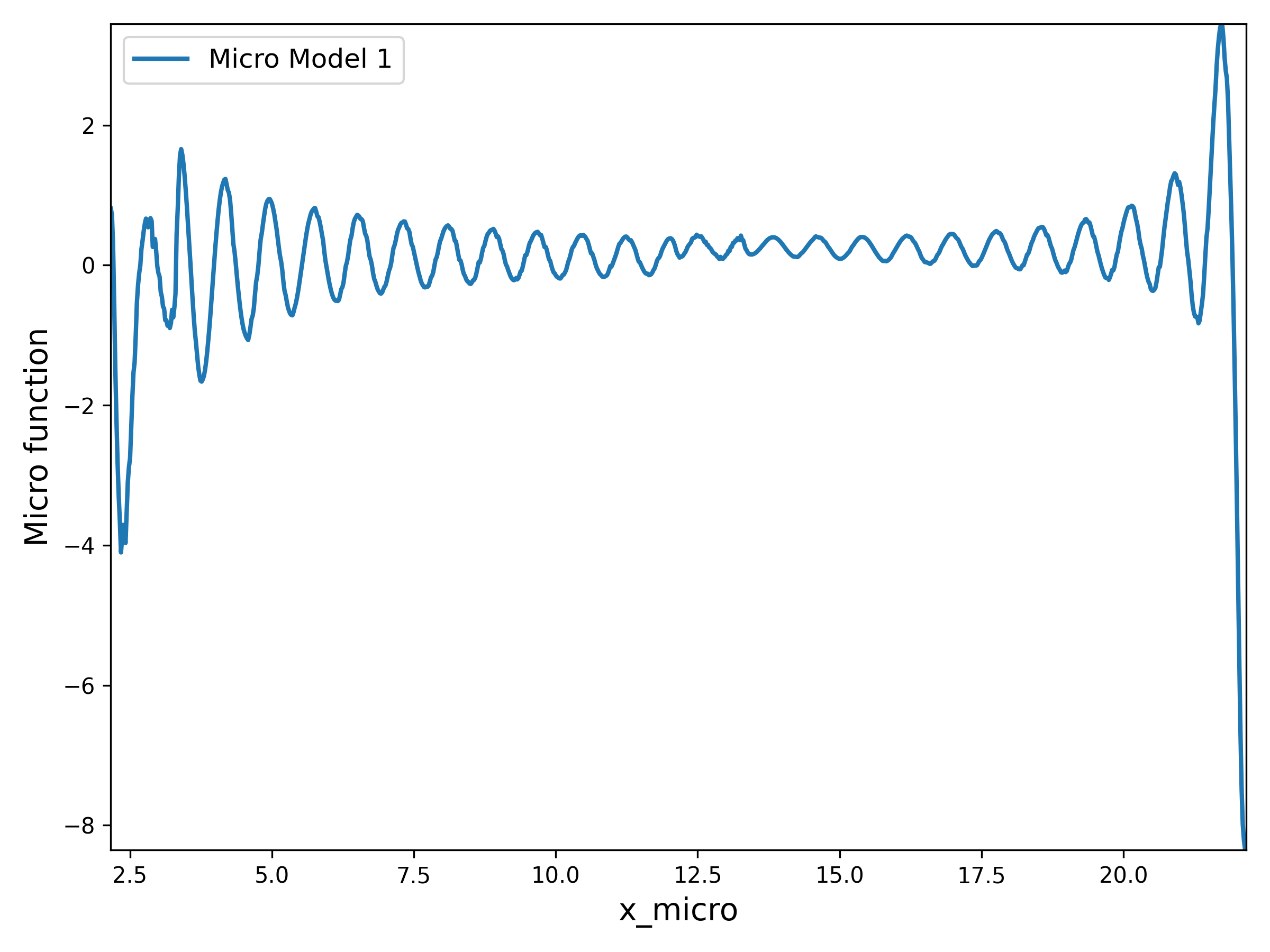}
            \caption{Micro component of Eq.~\eqref{eq:sin&exp}.}
            \label{fig:micro1}
        \end{subfigure}
    }

    \caption{PU approximation for Eq.~\eqref{eq:sin&exp}.}
    \label{fig:pu_first_example}
\end{figure}

We start with a simple example, given by the following function:
\begin{equation} \label{eq:sin&exp}
f(x) = \sin(8x) + 5 \exp(-0.2 x) + 1.
\end{equation}
This system involves a combination of sinusoidal and exponential terms. For this simpler system, it is sufficient to use only one mode to achieve accurate predictions, as the resulting mean squared error (MSE) is 0.008, which lies below the predefined threshold of \(10^{-2}\). By starting with such a simple example, we can build an understanding of the method and the foundation for more complex systems. The plots for the function, as well as the macro and micro components, are illustrated in Figure~\ref{fig:pu_first_example}.

\subsubsection{A second example}
In this section, we consider a more complex example described by the following function:
\begin{equation} \label{eq:cos&sin}
f(x) = A \left( \frac{\sin\left(\frac{1}{3} x\right) + \cos\left(\frac{2}{3} x\right) + \exp(-x^2) + \text{const}}{c} \right) - k \cdot x.
\end{equation}

This example introduces more intricate dynamics, combining two sinusoidal functions at different frequencies with exponential terms and a linear component. The system exhibits more complex behavior, requiring the use of a multi-modes approximation to capture both the global and fine-scale dynamics.

\begin{figure}[h!]
    \centering

    \begin{subfigure}{0.7\textwidth}
        \centering
        \includegraphics[width=\textwidth]{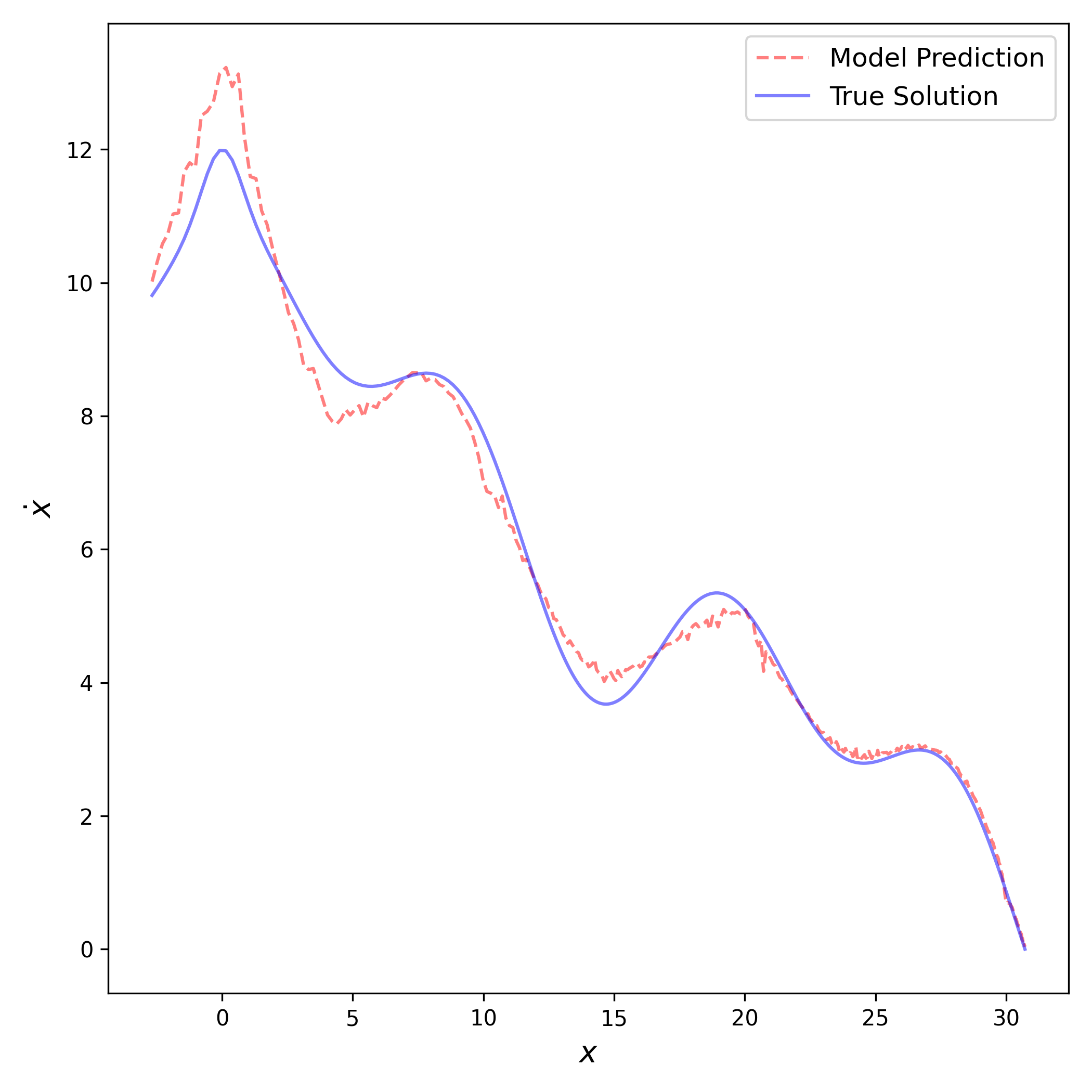}
        \caption{Function plot using one mode for Eq.~\eqref{eq:cos&sin}.}
        \label{fig:function2}
    \end{subfigure}
    
    \vspace{0.5cm}

    \makebox[\textwidth][c]{
        \begin{subfigure}{0.48\textwidth}
            \centering
            \includegraphics[width=\textwidth]{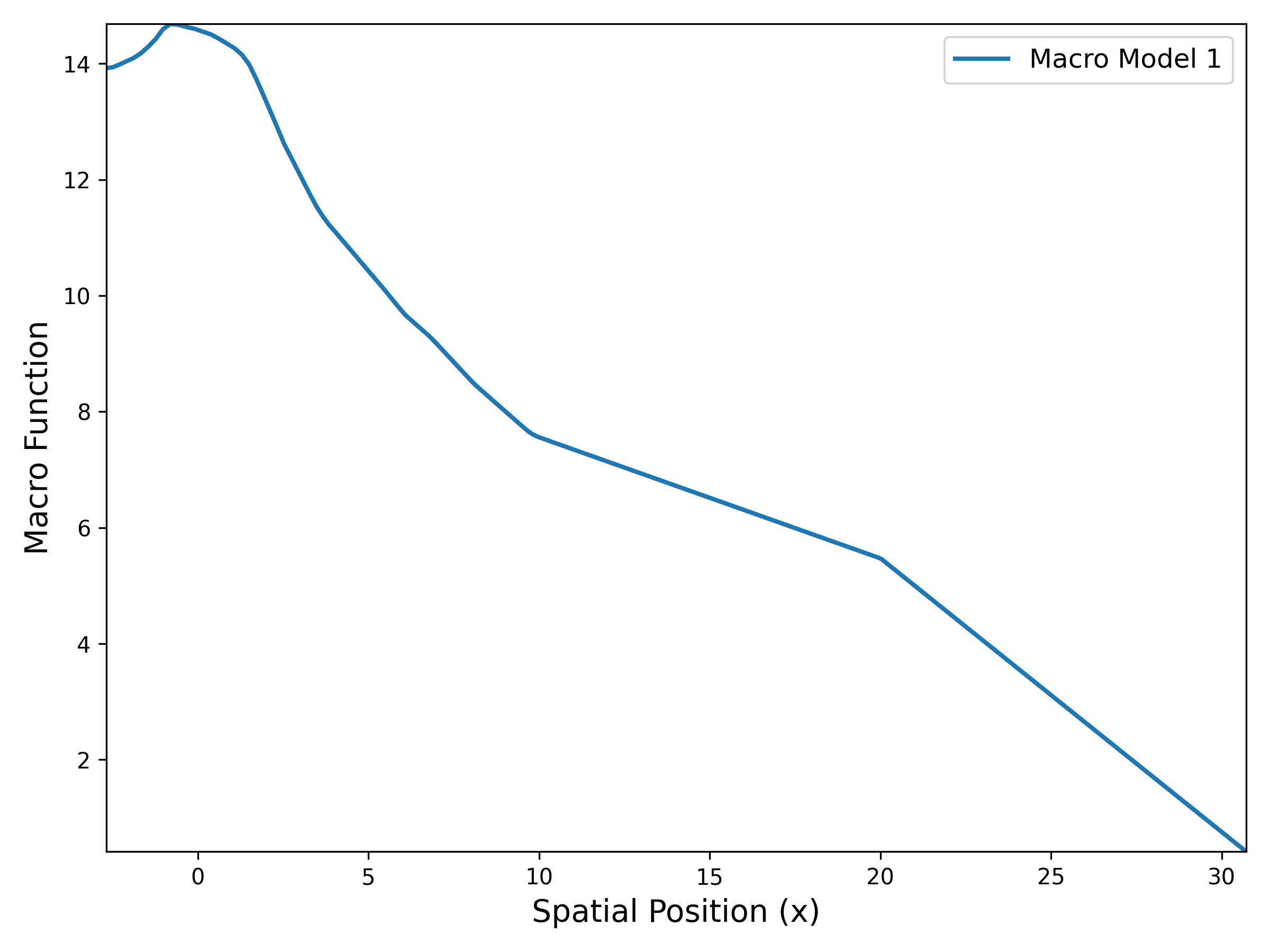}
            \caption{Macro component using one mode for Eq.~\eqref{eq:cos&sin}.}
            \label{fig:macro2}
        \end{subfigure}
        \hspace{0.04\textwidth}
        \begin{subfigure}{0.48\textwidth}
            \centering
            \includegraphics[width=\textwidth]{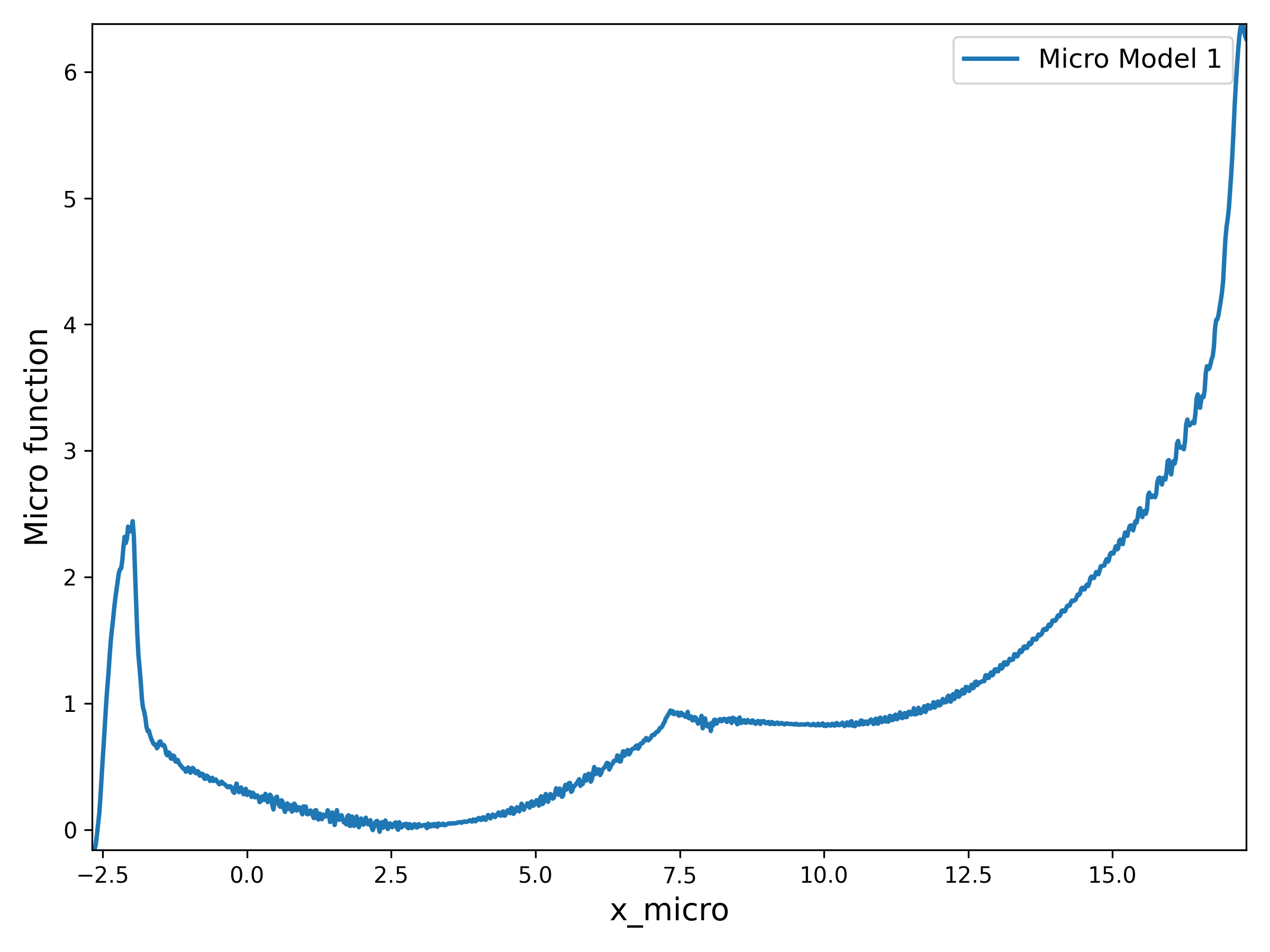}
            \caption{Micro component using one mode for Eq.~\eqref{eq:cos&sin}.}
            \label{fig:micro2}
        \end{subfigure}
    }

    \caption{PU approximation for Eq.~\eqref{eq:cos&sin} using one mode.}
    \label{fig:pu_second_example}
\end{figure}

We first show the results obtained using a single mode, which clearly illustrates that the model does not exhibit a good fitting. The mean squared error (MSE) in this case is 0.031, which is above the predefined threshold value, confirming that a single mode is insufficient to capture the full behavior of the system. Figure~\ref{fig:pu_second_example} shows the overall function along with its corresponding macro and micro components. These results highlight that the model does not fit well with only one mode, which is expected since the signal contains two distinct frequencies.

However, in this two-mode setup, the models are trained sequentially, in a parsimonious manner. The sequential training is handled using a wrapper module that allows multiple PU models to be combined and executed in sequence, ensuring that each mode is trained independently before composing the final model.

\begin{figure}[h!]
    \centering

    \begin{subfigure}{0.7\textwidth}
        \centering
        \includegraphics[width=\textwidth]{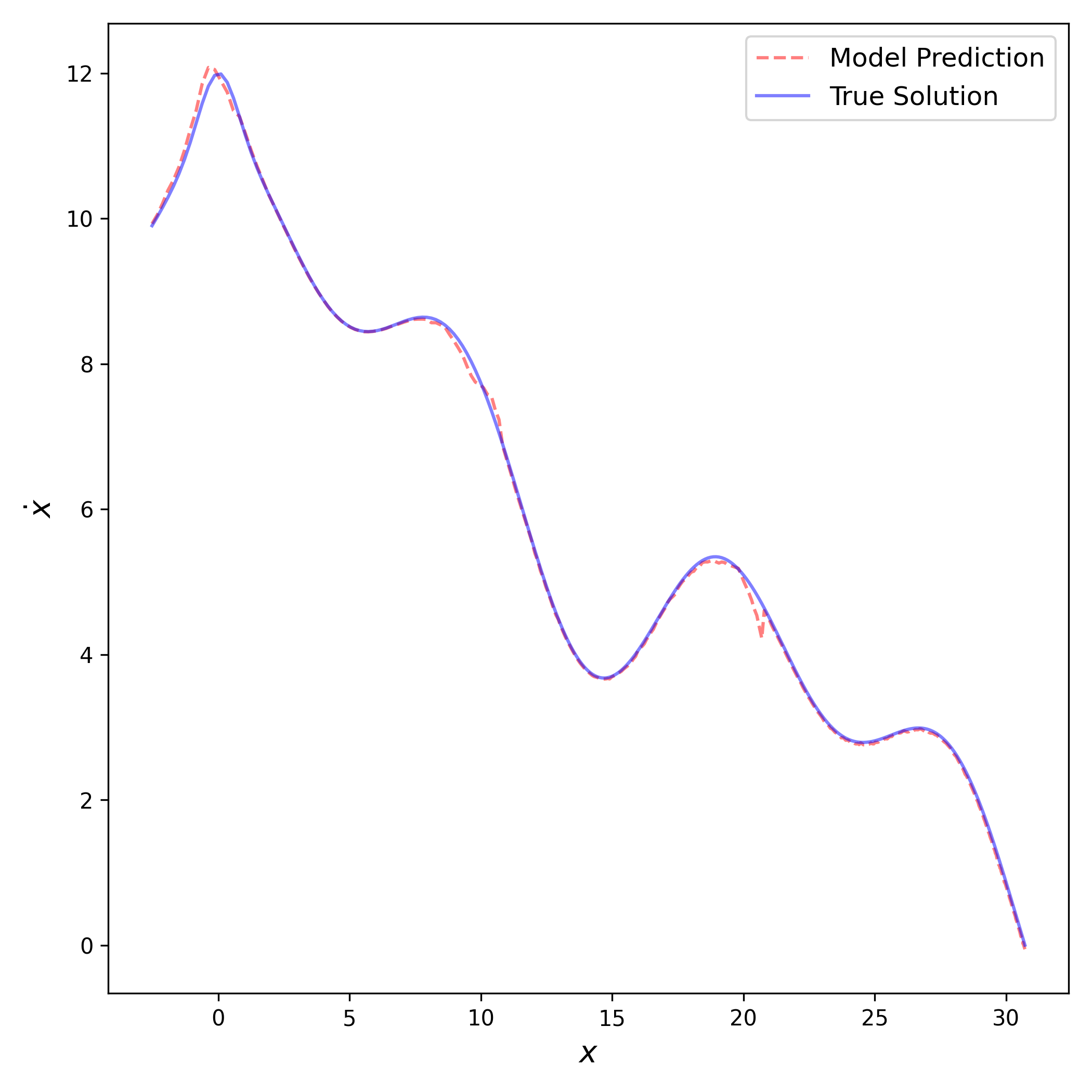}
        \caption{Function plot using two modes for Eq.~\eqref{eq:cos&sin}.}
        \label{fig:function3}
    \end{subfigure}
    
    \vspace{0.5cm}

    \makebox[\textwidth][c]{
        \begin{subfigure}{0.48\textwidth}
            \centering
            \includegraphics[width=\textwidth]{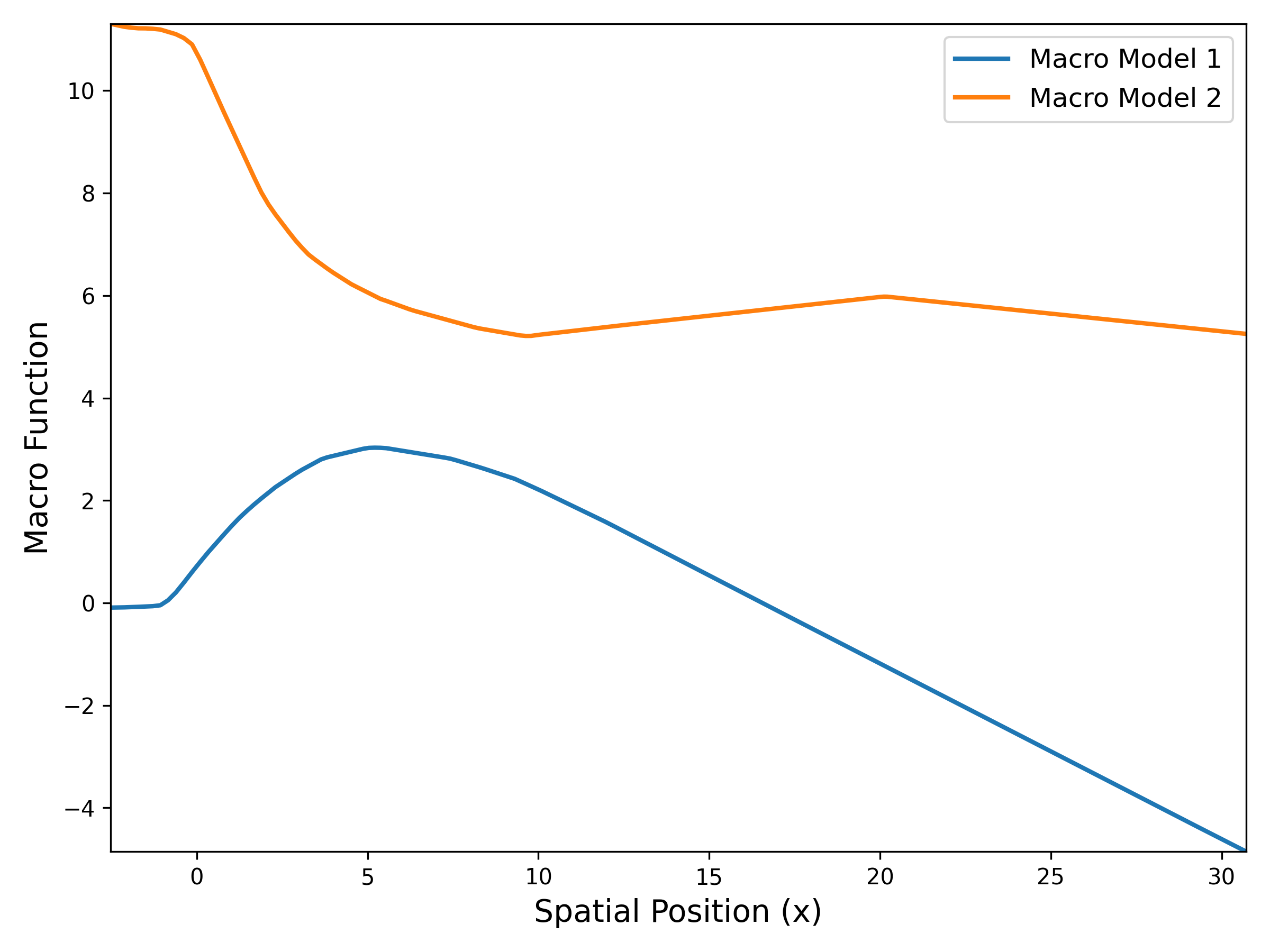}
            \caption{Macro component using two modes for Eq.~\eqref{eq:cos&sin}.}
            \label{fig:macro3}
        \end{subfigure}
        \hspace{0.04\textwidth}
        \begin{subfigure}{0.48\textwidth}
            \centering
            \includegraphics[width=\textwidth]{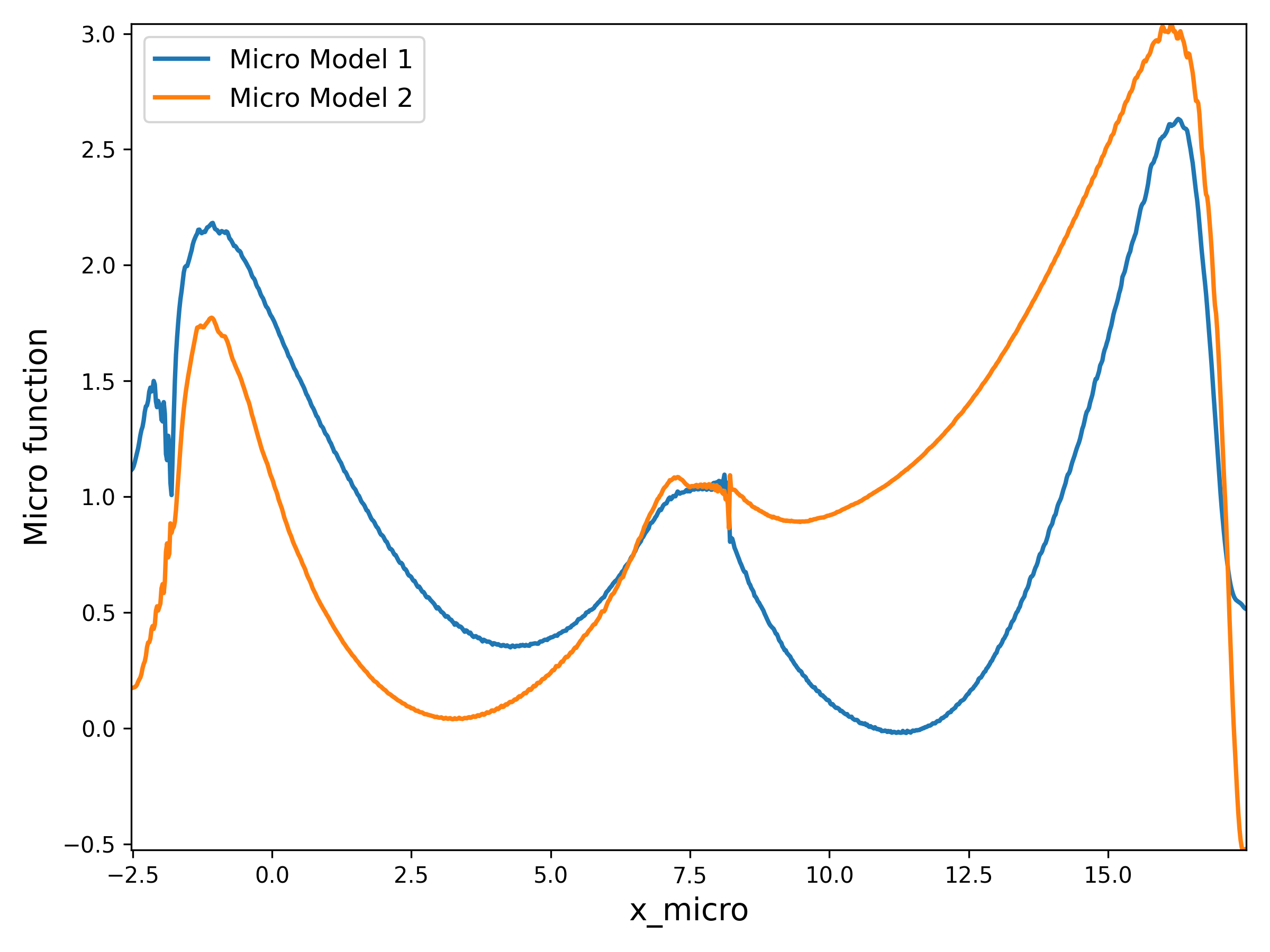}
            \caption{Micro component using two modes for Eq.~\eqref{eq:cos&sin}.}
            \label{fig:micro3}
        \end{subfigure}
    }

    \caption{PU approximation for Eq.~\eqref{eq:cos&sin} using two modes.}
    \label{fig:pu_second_example_2}
\end{figure}

By using two modes, the results improve significantly, providing a much better fit to both the global and fine-scale dynamics. The model achieves an MSE of 0.004, confirming the accuracy and adequacy of the two-mode configuration. Figure~\ref{fig:pu_second_example_2} presents the improved function along with its corresponding macro and micro components. The addition of the second mode significantly enhances the fit, capturing both the large-scale and fine-scale dynamics more accurately.

\subsubsection{Coupled systems}

\paragraph{Duffing System}

The Duffing system is governed by the following differential equations:
\begin{equation} \label{eq:duffin}
\frac{d}{dt} \begin{bmatrix} x_0 \\ x_1 \end{bmatrix} = \begin{bmatrix} x_1 \\ x_0 - x_0^3 \end{bmatrix}
\end{equation}

Both variables \(x_0\) and \(x_1\) interact nonlinearly. To simplify learning, we assume \(\dot{x}_1\) depends on \(x_0\) and \(\dot{x}_0\) depends on \(x_1\), allowing the system to be split into two separate 1D problems. Two Partition of Unity (PU) models were trained: one predicts \(\dot{x}_1\) from \(x_0\), and the other predicts \(\dot{x}_0\) from \(x_1\).

\begin{figure}[h!]
    \centering

    \begin{subfigure}{0.7\textwidth}
        \centering
        \includegraphics[width=\textwidth]{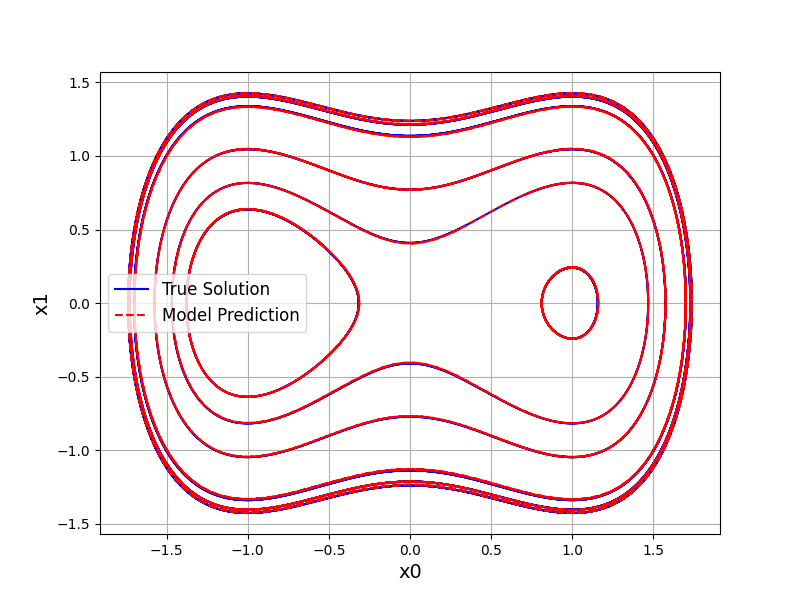}
        \caption{Function plot for Eq.~\eqref{eq:duffin}.}
        \label{fig:duffin}
    \end{subfigure}

    \vspace{0.5cm}

    \makebox[\textwidth][c]{
        \begin{subfigure}{0.48\textwidth}
            \centering
            \includegraphics[width=\textwidth]{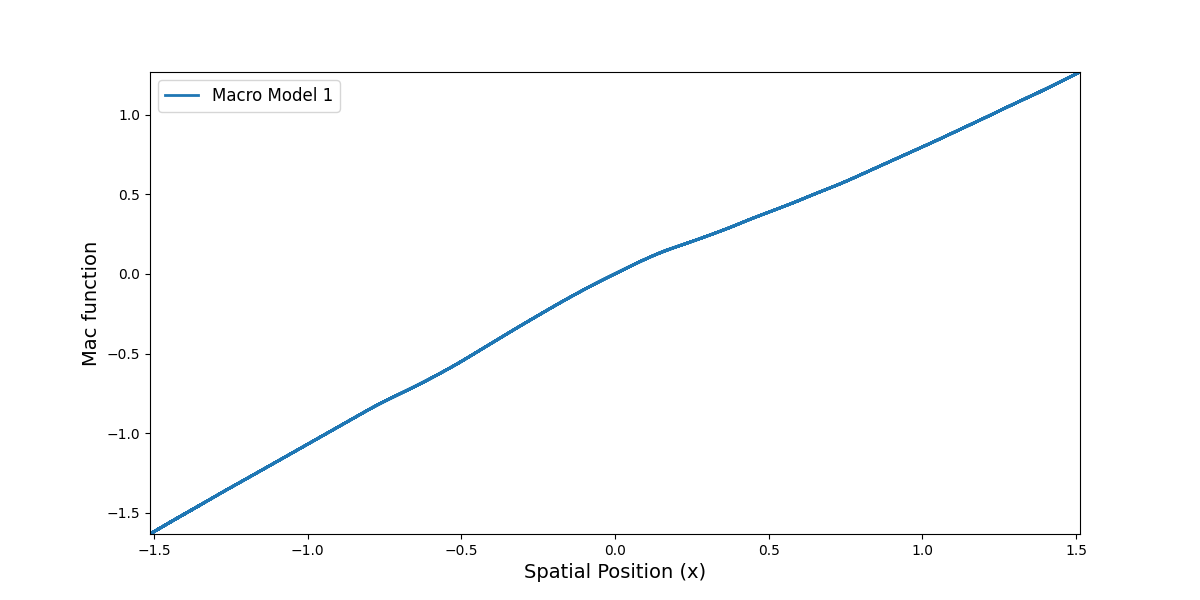}
            \caption{Macro component 1 for Eq.~\eqref{eq:duffin}.}
            \label{fig:macrod1}
        \end{subfigure}
        \hspace{0.04\textwidth}
        \begin{subfigure}{0.48\textwidth}
            \centering
            \includegraphics[width=\textwidth]{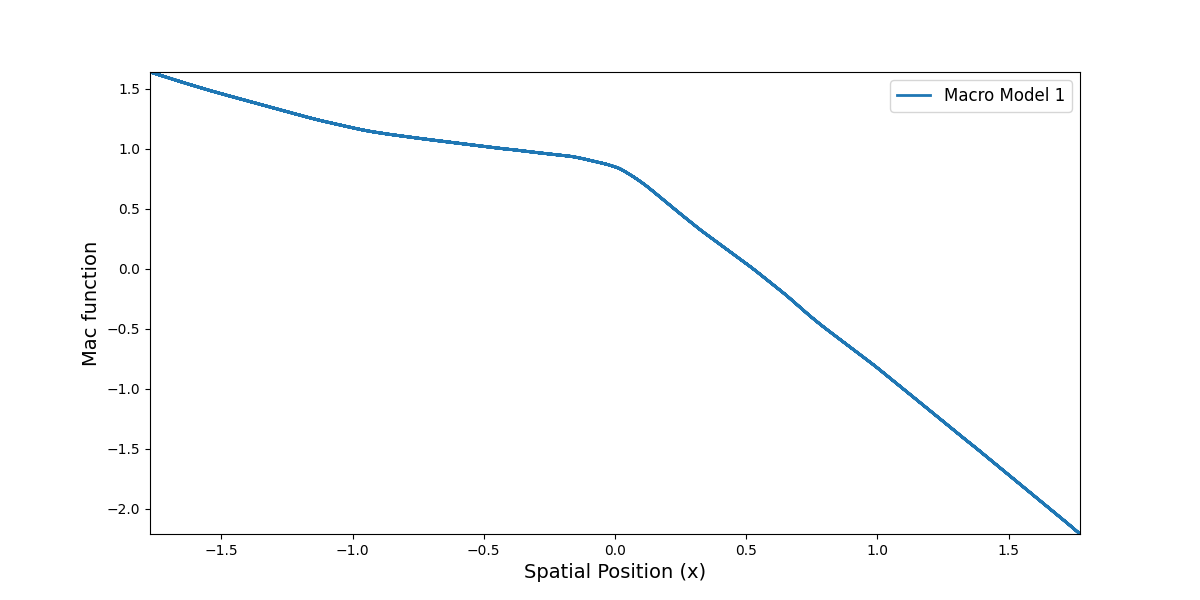}
            \caption{Macro component 2 for Eq.~\eqref{eq:duffin}.}
            \label{fig:macrod2}
        \end{subfigure}
    }

    \vspace{0.5cm}

    \makebox[\textwidth][c]{
        \begin{subfigure}{0.48\textwidth}
            \centering
            \includegraphics[width=\textwidth]{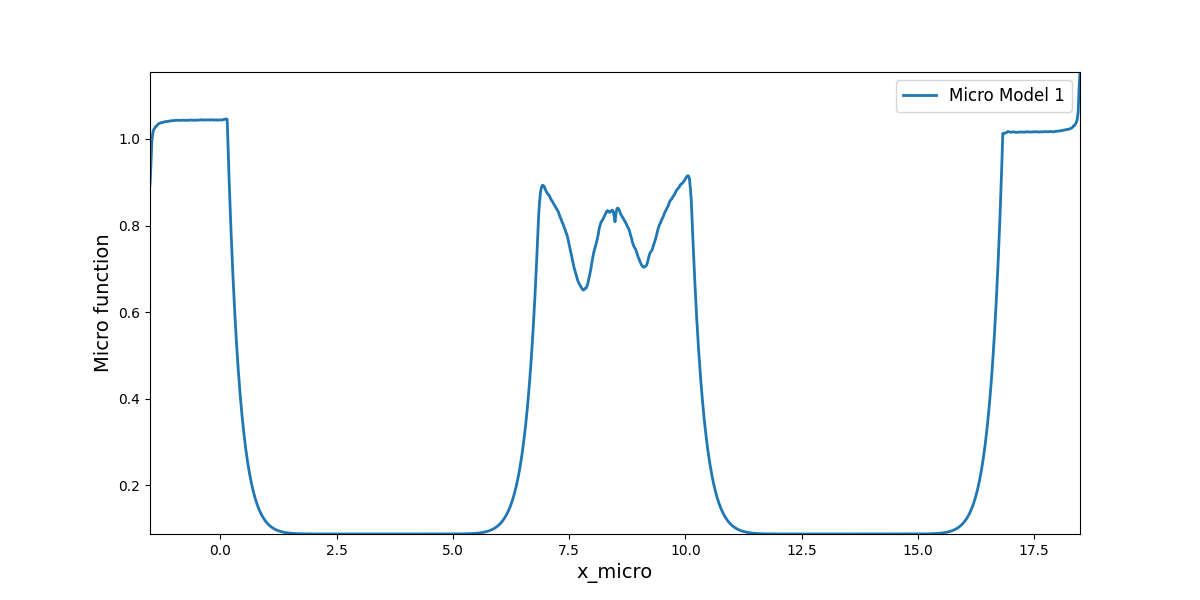}
            \caption{Micro component 1 for Eq.~\eqref{eq:duffin}.}
            \label{fig:microd1}
        \end{subfigure}
        \hspace{0.04\textwidth}
        \begin{subfigure}{0.48\textwidth}
            \centering
            \includegraphics[width=\textwidth]{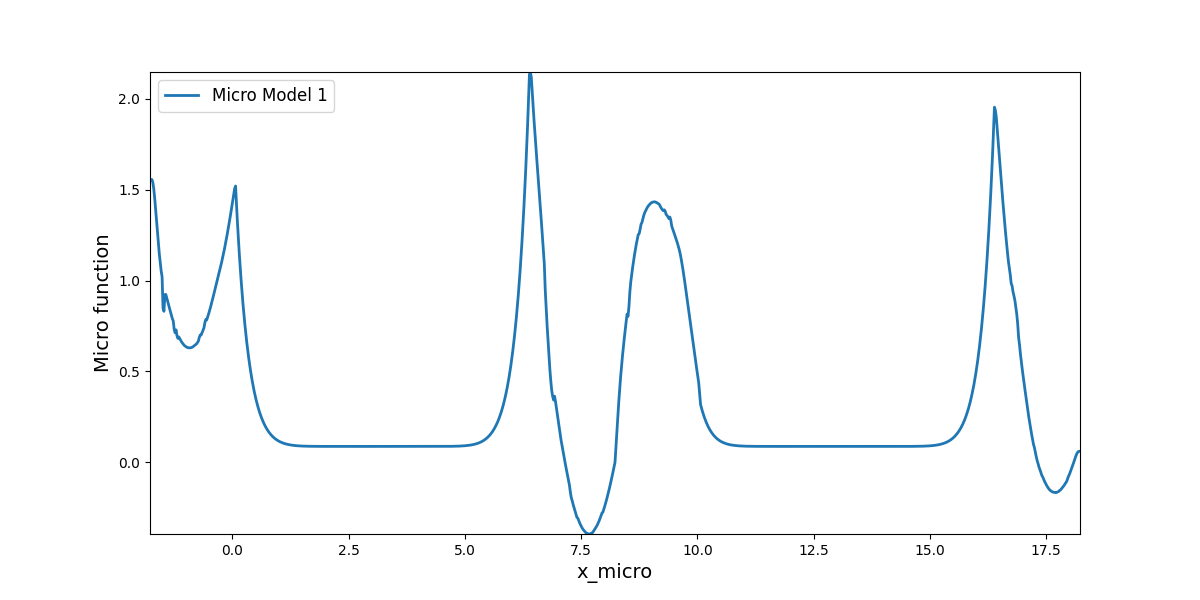}
            \caption{Micro component 2 for Eq.~\eqref{eq:duffin}.}
            \label{fig:microd2}
        \end{subfigure}
    }

    \caption{PU approximation for Eq.~\eqref{eq:duffin}.}
    \label{fig:pu_duffin_all}
\end{figure}

The results demonstrate that the PU method effectively captures the dynamics of the Duffing system, achieving a mean squared error of 0.001. Figure~\ref{fig:pu_duffin_all} shows the function and its corresponding macro and micro components, providing accurate predictions of the system behavior.

\paragraph{Harmonic Oscillator System}

The harmonic oscillator system is described by:
\begin{equation} \label{eq:harmonic}
\frac{d}{dt} \begin{bmatrix} x_0 \\ x_1 \end{bmatrix} = \begin{bmatrix} x_1 \\ -\sin(x_0) \end{bmatrix}
\end{equation}

Similarly, we train two PU models for the oscillator: one predicting \(\dot{x}_1\) from \(x_0\) and the other predicting \(\dot{x}_0\) from \(x_1\), yielding two macro and micro components.

\begin{figure}[h!]
    \centering

    \begin{subfigure}{0.7\textwidth}
        \centering
        \includegraphics[width=\textwidth]{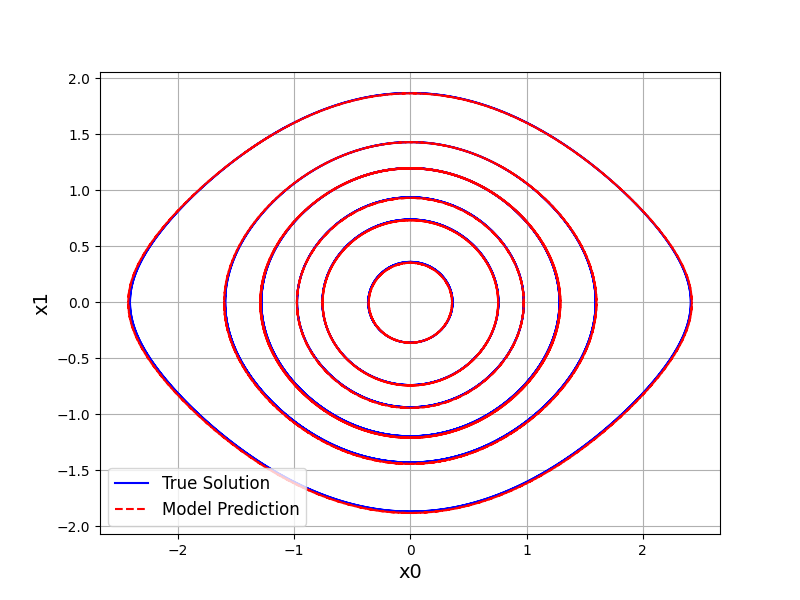}
        \caption{Function plot for Eq.~\eqref{eq:harmonic}.}
        \label{fig:harmonic}
    \end{subfigure}

    \vspace{0.5cm}

    \makebox[\textwidth][c]{
        \begin{subfigure}{0.48\textwidth}
            \centering
            \includegraphics[width=\textwidth]{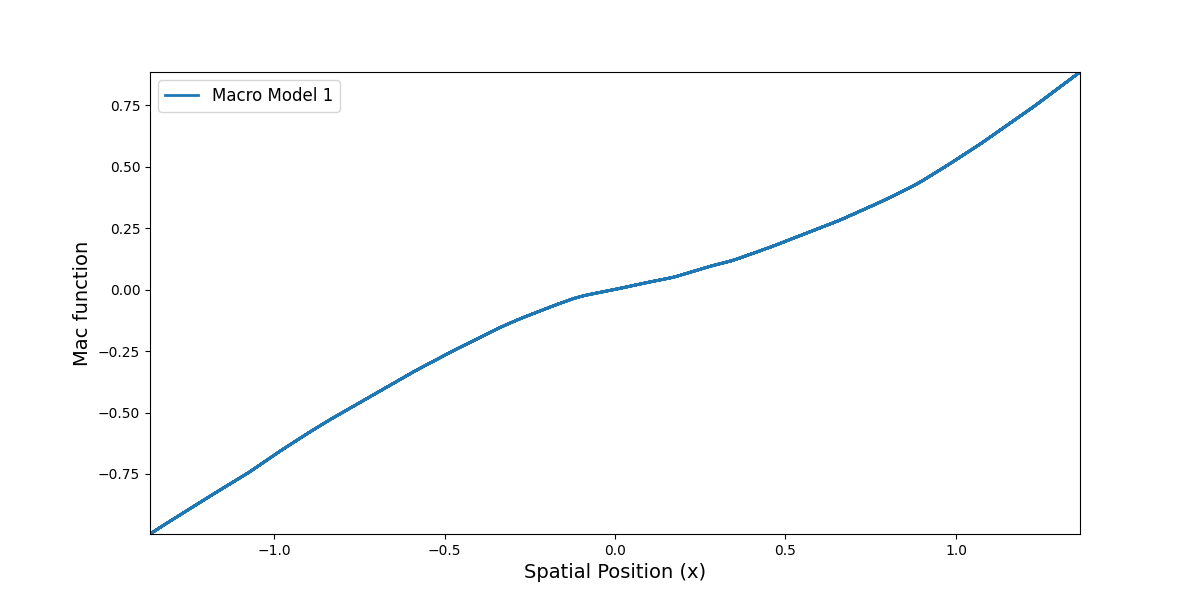}
            \caption{Macro component 1 for Eq.~\eqref{eq:harmonic}.}
            \label{fig:macroh1}
        \end{subfigure}
        \hspace{0.04\textwidth}
        \begin{subfigure}{0.48\textwidth}
            \centering
            \includegraphics[width=\textwidth]{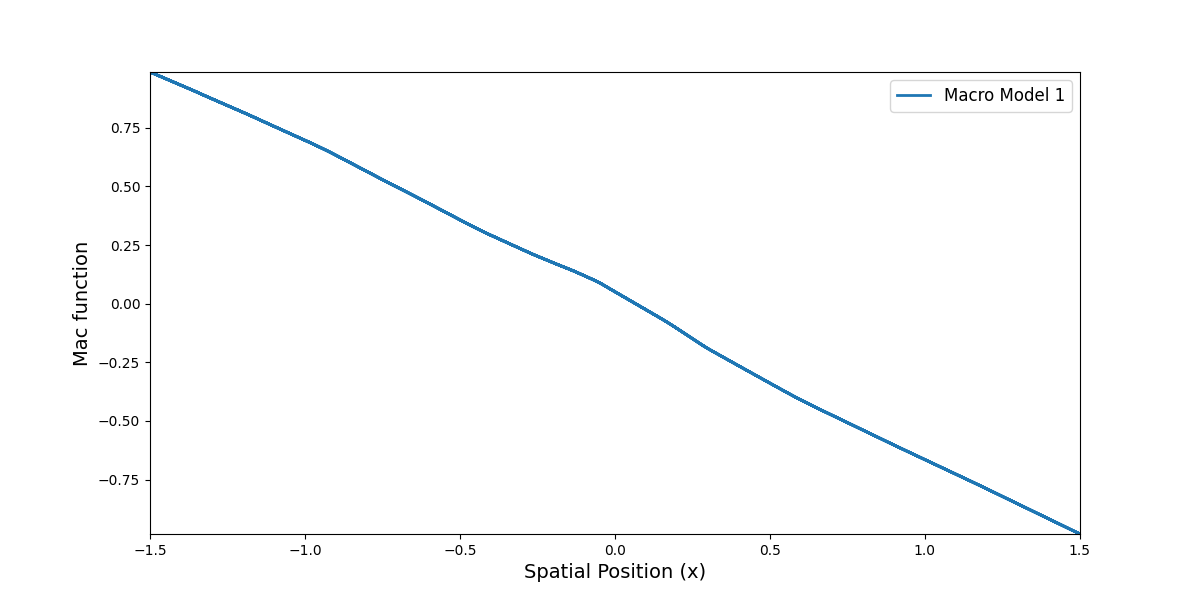}
            \caption{Macro component 2 for Eq.~\eqref{eq:harmonic}.}
            \label{fig:macroh2}
        \end{subfigure}
    }

    \vspace{0.5cm}

    \makebox[\textwidth][c]{
        \begin{subfigure}{0.48\textwidth}
            \centering
            \includegraphics[width=\textwidth]{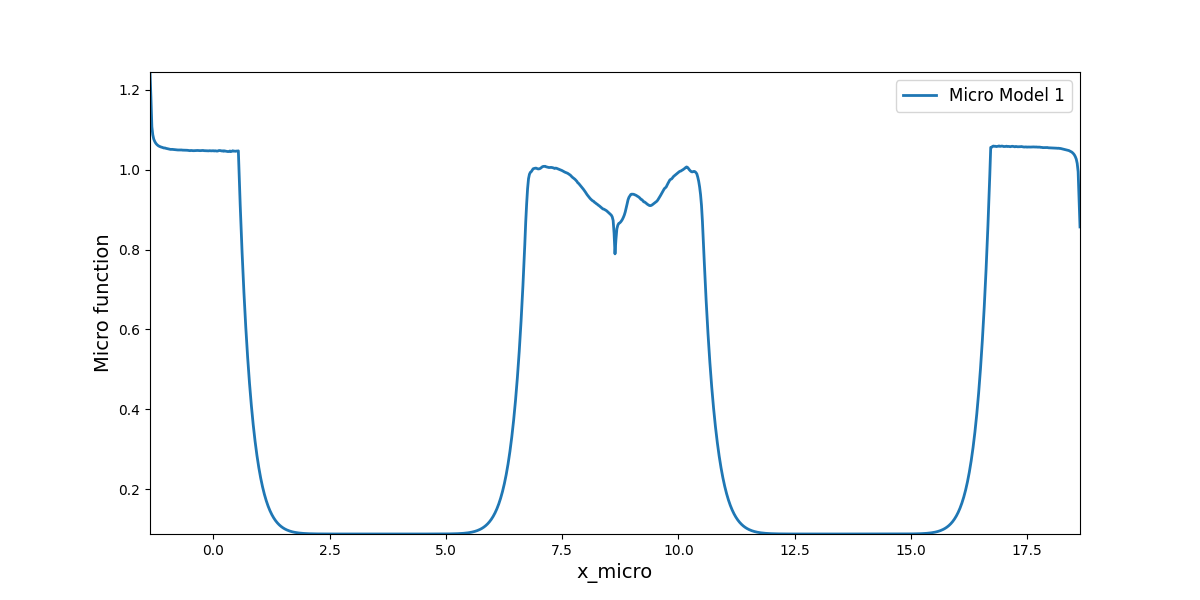}
            \caption{Micro component 1 for Eq.~\eqref{eq:harmonic}.}
            \label{fig:microh1}
        \end{subfigure}
        \hspace{0.04\textwidth}
        \begin{subfigure}{0.48\textwidth}
            \centering
            \includegraphics[width=\textwidth]{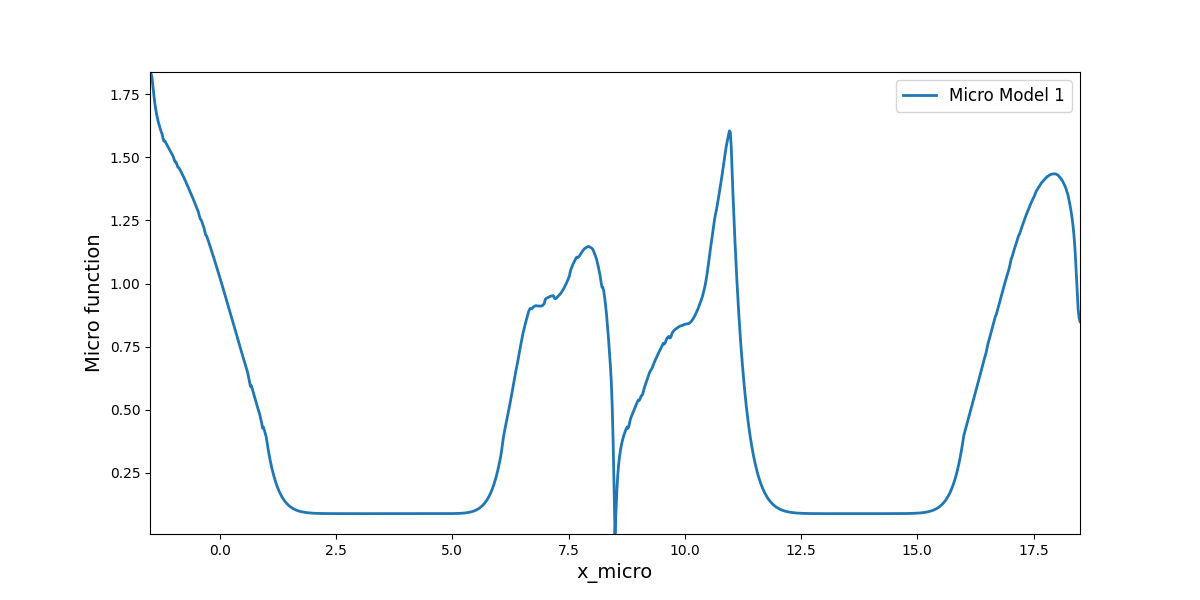}
            \caption{Micro component 2 for Eq.~\eqref{eq:harmonic}.}
            \label{fig:microh2}
        \end{subfigure}
    }

    \caption{PU approximation for Eq.~\eqref{eq:harmonic}.}
    \label{fig:pu_harmonic_all}
\end{figure}

The results show that the PU method successfully models the harmonic oscillator, achieving a mean squared error (MSE) of 0.0006. Figure~\ref{fig:pu_harmonic_all} shows the function and its corresponding macro and micro components, capturing both coarse and fine-scale dynamics with high accuracy and efficiency.

\subsubsection{Summary of PU Results}
In summary, the numerical results for all the considered systems demonstrate that the PU method is an effective tool for modeling and predicting complex, multi-scale dynamical systems. By separating each system’s dynamics into coarse and fine scales, the method efficiently captures both global trends and detailed fluctuations. This provides a comprehensive view of the system's behavior with high accuracy and computational efficiency, as illustrated by the results for the Duffing and harmonic oscillator systems.

\section{Singular Value Decomposition (SVD) for multi-scale function extraction}

\begin{figure}[h!]
\centering
\begin{tikzpicture}[>=Stealth, every node/.style={align=center}]

\node (plot) at (0,5) {
    \begin{tikzpicture}[scale=0.8]
        \draw[->] (0,0) -- (10.5,0) node[right] {$x$};
        \draw[->] (0,0) -- (0,3) node[above] {$f(x)$};
        \draw[thick, domain=0:10.2, samples=100, smooth] plot (\x,{1 + 0.5* sin(3*\x r)});
        \draw[blue, dashed] (2.5,0) -- (2.5,2) ;
        \draw[blue, dashed] (5,0) -- (5,2) ;
        \draw[blue, dashed] (7.5,0) -- (7.5,2) ;
        \draw[blue, dashed] (10,0) -- (10,2) ;
        
        \foreach \i/\label in {0.1/{$x_{11}$}, 1/{$x_{12}$}, 2.45/{$x_{1n}$}} {
        \fill[red] (\i,0) circle (1.5pt);
        \node[below] at (\i,0) {\label};
        }
        
        \foreach \i/\label in {7.6/{$x_{m1}$}, 8.5/{$x_{m2}$}, 9.95/{$x_{mn}$}} {
        \fill[red] (\i,0) circle (1.5pt);
        \node[below] at (\i,0) {\label};
        }

    \end{tikzpicture}
};

\draw[->, thick] (0,3.5) -- (0,2.2);

\node[draw, thick, rounded corners, fill=gray!15, minimum width=6cm, minimum height=1.8cm] (matrix) at (0,1) {
    $\boldsymbol{F} = \begin{bmatrix}
    x_{11} & x_{12} & \cdots & x_{1n} \\
    x_{21} & x_{22} & \cdots & x_{2n} \\
    \vdots & \vdots & \ddots & \vdots \\
    x_{m1} & x_{m2} & \cdots & x_{mn}
    \end{bmatrix}$};

\draw[->, thick] (0,-0.3) -- (0,-1.2);

\node[draw, thick, rounded corners, fill=green!15, minimum width=6cm, minimum height=1.2cm] (svd) at (0,-2) {$\boldsymbol{F} \approx \boldsymbol{U} \boldsymbol{\Sigma} \boldsymbol{V}^T$ \\ {\scriptsize $\boldsymbol{U}$: macro, $\boldsymbol{\Sigma} \boldsymbol{V}^T$: micro}};

\end{tikzpicture}
\caption{Illustration of the SVD workflow. The original function is divided into segments, transformed into a matrix $\boldsymbol{F}$, and then decomposed using SVD. The left singular vectors $\boldsymbol{U}$ capture the macro-scale structure, while $\boldsymbol{\Sigma} \boldsymbol{V}^T$ captures fine-scale variations (micro).}
\label{fig:svd_macro_micro}
\end{figure}
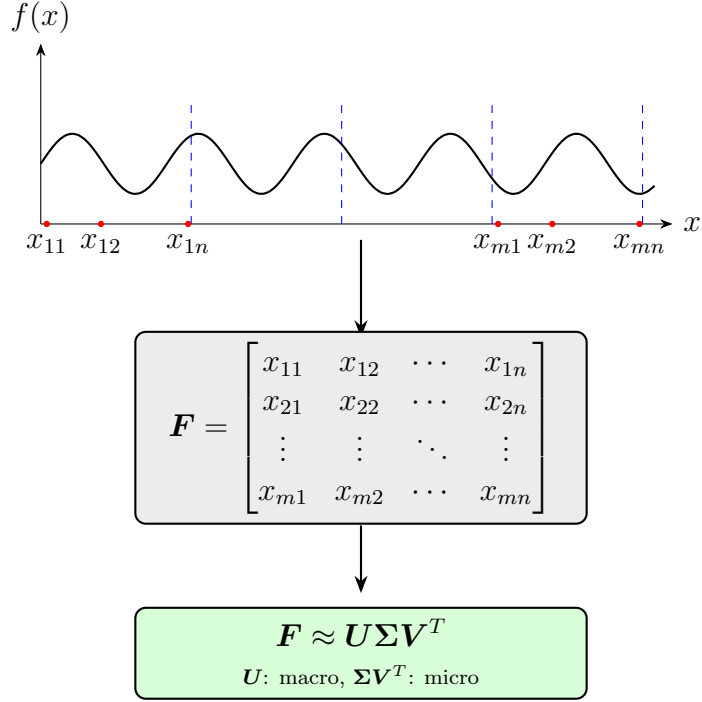

The Partition of Unity (PU) method constructs local approximations of the system’s dynamics, capturing both macro- and micro-scale behaviors. Another method for analyzing the system is the singular value decomposition (SVD). By applying SVD to the data, the extracted modes can similarly be interpreted in terms of macro- and micro-scale components, providing a connection to the scale separation achieved by the PU method.

To perform the SVD, one could construct the data matrix by grouping overlapping macro-elements, i.e., combining each macro-element with its neighboring ones, similar to the overlapping approach used in the PU method. We instead use non-overlapping macro-elements. This choice avoids introducing strong linear dependencies between blocks, ensuring that the SVD extracts independent modes that represent the macro- and micro-scale structure of the system.

\subsection{Domain Discretization}

In this section, we assume that the \(x\)-domain is equipped with a fine mesh that can capture all the scales involved in the dynamics, consisting of \(C\) nodes. A coarse mesh is then constructed, consisting of \(m\) macro-elements, each of which contains \(n\) micro-nodes, such that
\[
C = m \times n.
\]
Thus, each original node \(x_{ij}\) can be indexed by a pair \((i,j)\), where 
\(i = 1, \dots, m\) denotes the macro-element index, and 
\(j = 1, \dots, n\) denotes the micro-node within the \(i\)-th macro-element.

The function nodal values \( f(x_{ij}) \), for all \(i, j\), can now be expressed as a matrix \( \boldsymbol{F} \), which has \( m \) rows and \( n \) columns. This representation does not introduce any complexity reduction at this stage, but it is well-known that matrices can be efficiently approximated in a hierarchical manner using the Singular Value Decomposition (SVD) (see Figure~\ref{fig:svd_macro_micro}).

\subsection{SVD Representation}
Instead of learning the matrix \( \boldsymbol{F} \) directly, it is more efficient to learn its SVD representation. The SVD allows the matrix to be expressed as the sum of outer products of singular vectors:

\[
\boldsymbol{F} \approx \sum_{i=1}^{T} \mathbf{U}_i \otimes \mathbf{V}_i,
\]

where \( \mathbf{U}_i \) and \( \mathbf{V}_i \) are the left and right singular vectors, respectively, and \( T \) is the number of modes retained in the truncated approximation.

The truncated SVD representation enables a significant reduction in complexity, as it allows to approximate \( \boldsymbol{F} \) using only the most significant singular values and vectors. By truncating the SVD to a limited number of modes \( T \), we can achieve a compact representation of the matrix \( \boldsymbol{F} \) that still captures the essential information needed to describe the system's dynamics. The matrix \( \boldsymbol{F} \) used to compute the SVD has the micro and macro scales represented in the construction of each row and column. As a result, the left (truncated) singular vectors computed with the SVD will represent the micro functions equivalently to the PU, and the right (truncated) singular vectors computed with the SVD will represent the macro functions.

This approach allows for a more efficient representation of the system, as it significantly reduces the number of parameters required to model the system's evolution. The computational benefits of using the SVD approximation are particularly evident when dealing with large-scale systems or systems with many degrees of freedom.

In summary, the SVD method offers a powerful way to approximate the nodal values of the system using a hierarchical decomposition that can be truncated to capture the essential features of the dynamics with reduced computational cost. This makes it an attractive technique for efficiently modeling complex systems with multi-scale behavior.

\subsection{Numerical results}

We now present the results obtained using the Singular Value Decomposition (SVD) method, applied to the same examples as the Partition of Unity (PU) method.

For each system, the SVD method was applied to approximate the nodal values \( f(x_{ij}) \) by decomposing the matrix \( \boldsymbol{F} \). In the application of SVD, the selection of modes is typically determined by the energy captured by the modes or by the reconstruction error, as this reflects the decay behavior of the singular values in the analysis. In this work, we adopted an error-based criterion, selecting the number of modes such that the mean squared error (MSE) remained below \(10^{-2}\), consistent with the criterion established in the PU method.

\begin{figure}[h!]
    \centering

    \begin{subfigure}{0.7\textwidth}
        \centering
        \includegraphics[width=\textwidth]{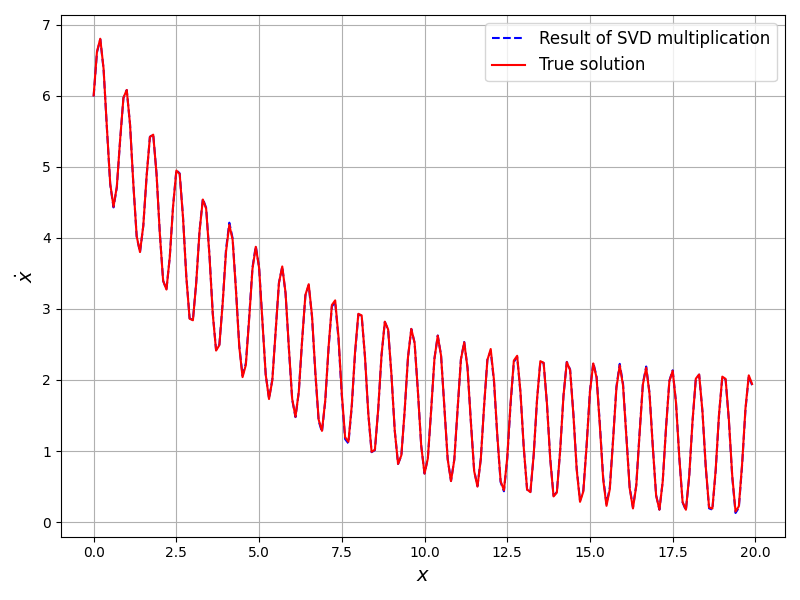}
        \caption{Function Plot using three modes for Eq.~\eqref{eq:sin&exp}.}
        \label{fig:sin&exp_svd3_plot}
    \end{subfigure}
    
    \vspace{0.5cm}
    
    \makebox[\textwidth][c]{
        \begin{subfigure}{0.48\textwidth}
            \centering
            \includegraphics[width=\textwidth]{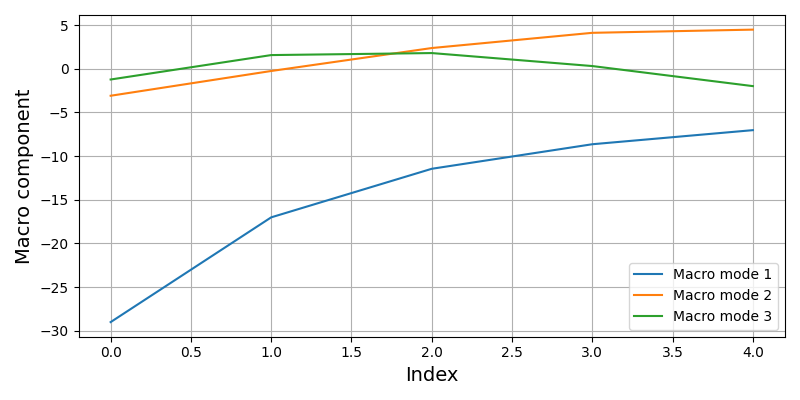}
            \caption{Macro component using three modes for Eq.~\eqref{eq:sin&exp}.}
            \label{fig:sin&exp_svd3_macro}
        \end{subfigure}
        \hspace{0.04\textwidth}
        \begin{subfigure}{0.48\textwidth}
            \centering
            \includegraphics[width=\textwidth]{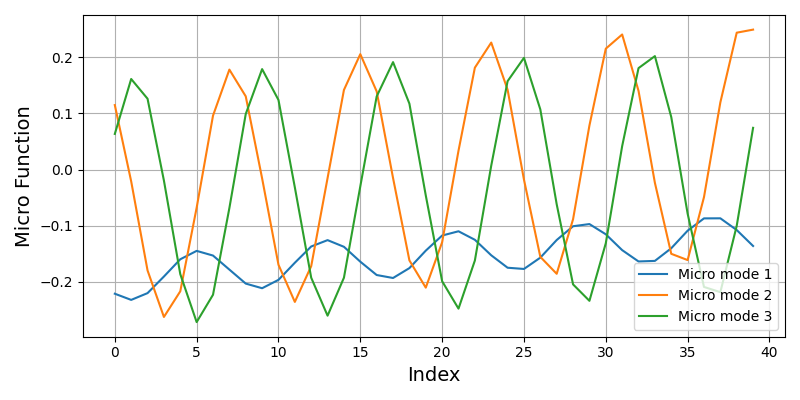}
            \caption{Micro component using three modes for Eq.~\eqref{eq:sin&exp}.}
            \label{fig:sin&exp_svd3_micro}
        \end{subfigure}
    }

    \caption{SVD approximation for Eq.~\eqref{eq:sin&exp} using three modes.}
    \label{fig:sin&exp_svd3_all}
\end{figure}

As illustrated in Section~\ref{firstexample}, we start with the equation~\eqref{eq:sin&exp}. As shown in Figure~\ref{fig:sin&exp_svd3_all}, the function and its corresponding macro and micro components are accurately represented when using three modes. The reconstruction achieved a mean squared error (MSE) of \(1.56 \times 10^{-4}\), which is below the predefined threshold criterion of \(10^{-2}\), confirming the adequacy of the selected number of modes.

\begin{figure}[h!]
    \centering

    \begin{subfigure}{0.7\textwidth}
        \centering
        \includegraphics[width=\textwidth]{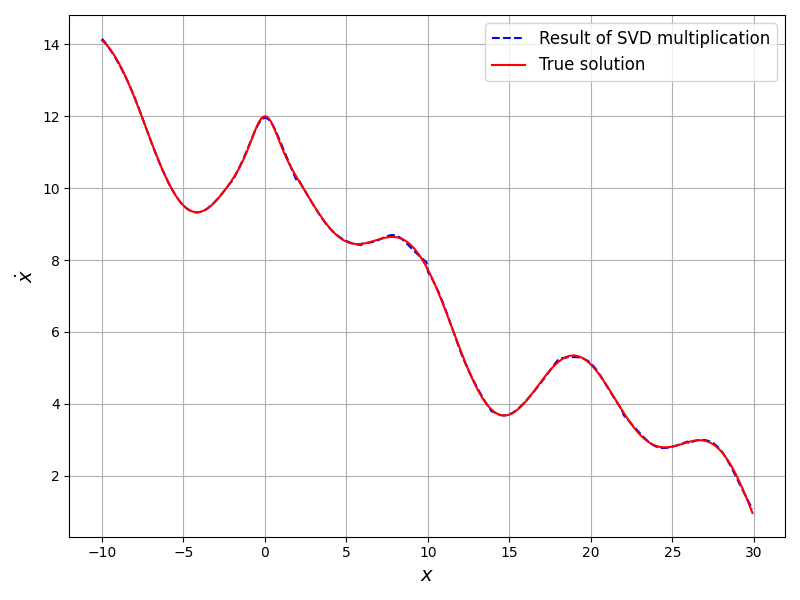}
        \caption{Function Plot using three modes for Eq.~\eqref{eq:cos&sin}.}
        \label{fig:cos&sin_svd3_plot}
    \end{subfigure}
    
    \vspace{0.5cm}
    
    \makebox[\textwidth][c]{
        \begin{subfigure}{0.48\textwidth}
            \centering
            \includegraphics[width=\textwidth]{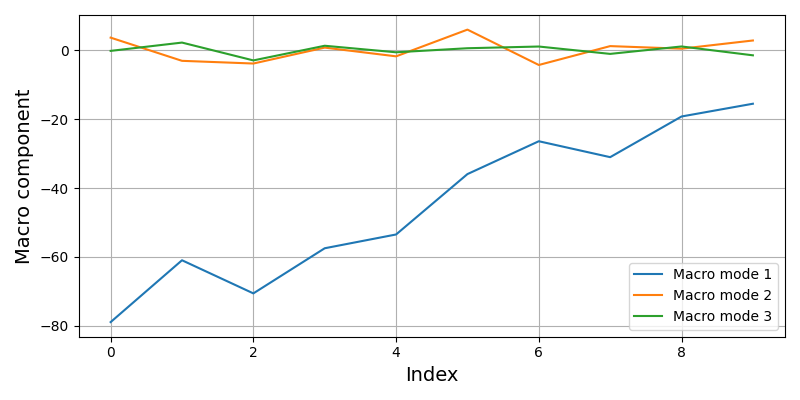}
            \caption{Macro component using three modes for Eq.~\eqref{eq:cos&sin}.}
            \label{fig:cos&sin_svd3_macro}
        \end{subfigure}
        \hspace{0.04\textwidth}
        \begin{subfigure}{0.48\textwidth}
            \centering
            \includegraphics[width=\textwidth]{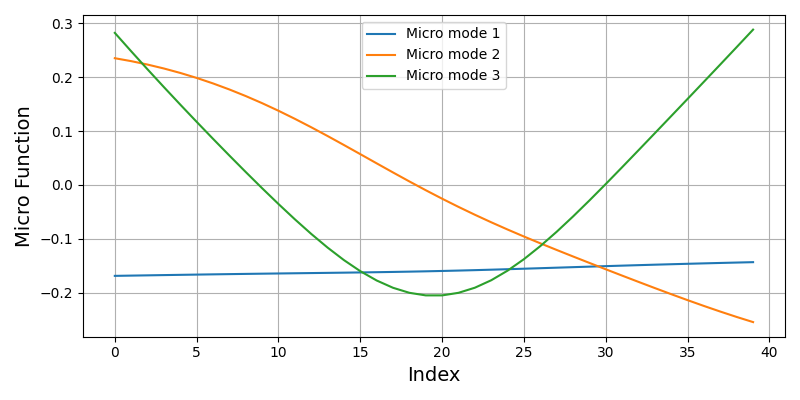}
            \caption{Micro component using three modes for Eq.~\eqref{eq:cos&sin}.}
            \label{fig:cos&sin_svd3_micro}
        \end{subfigure}
    }

    \caption{SVD approximation for Eq.~\eqref{eq:cos&sin} using three modes.}
    \label{fig:cos&sin_svd3_all}
\end{figure}

Next, we consider the dynamical system described by~\eqref{eq:cos&sin}. As illustrated in Figure~\ref{fig:cos&sin_svd3_all}, the prediction obtained using three modes provides an accurate approximation of the system dynamics. The function plot, together with its corresponding macro and micro components, demonstrates that the SVD method effectively captures both the global and local features of the system. The reconstruction achieved a mean squared error (MSE) of \(7.33 \times 10^{-4}\), which remains well below the predefined threshold, confirming the suitability of the selected mode configuration.

\begin{figure}[h!]
    \centering

    \begin{subfigure}{0.7\textwidth}
        \centering
        \includegraphics[width=\textwidth]{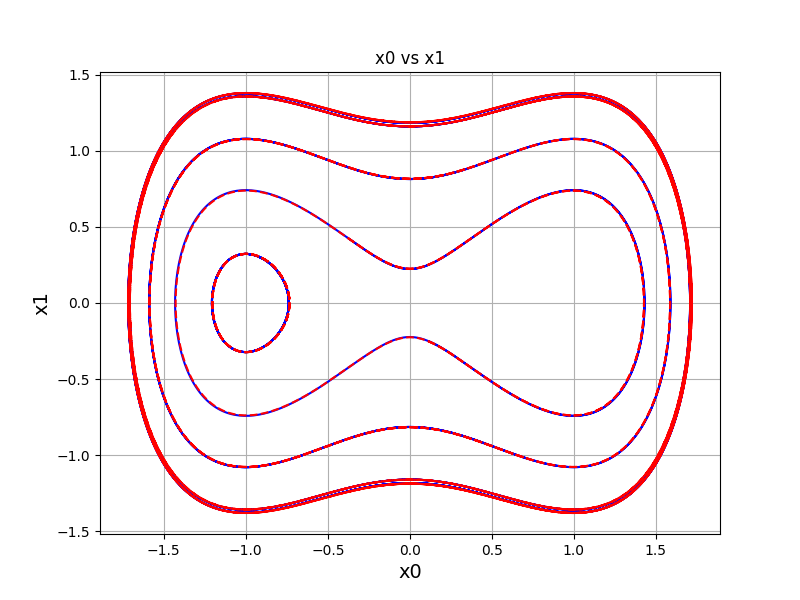}
        \caption{Function plot using three modes for Eq.~\eqref{eq:duffin}.}
        \label{fig:duffin_svd3_function}
    \end{subfigure}

    \vspace{0.5cm}

    \makebox[\textwidth][c]{
        \begin{subfigure}{0.48\textwidth}
            \centering
            \includegraphics[width=\textwidth]{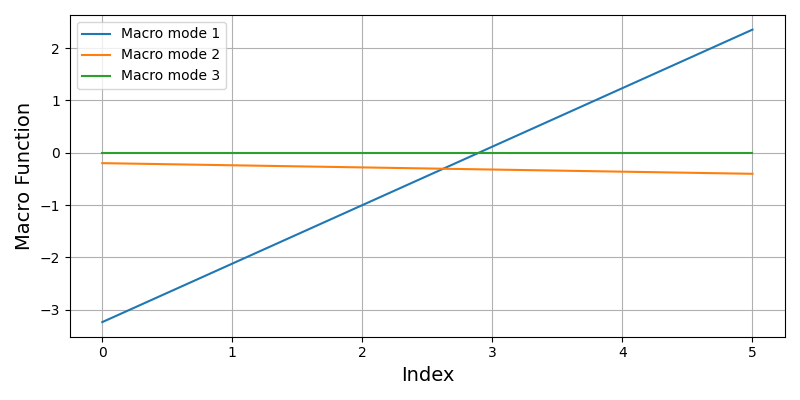}
            \caption{Macro component 1 for Eq.~\eqref{eq:duffin}.}
            \label{fig:duffin_svd3_macro1}
        \end{subfigure}
        \hspace{0.04\textwidth}
        \begin{subfigure}{0.48\textwidth}
            \centering
            \includegraphics[width=\textwidth]{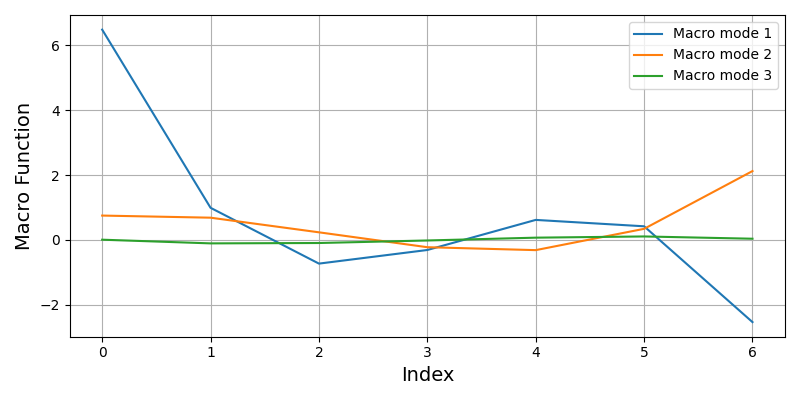}
            \caption{Macro component 2 for Eq.~\eqref{eq:duffin}.}
            \label{fig:duffin_svd3_macro2}
        \end{subfigure}
    }

    \vspace{0.5cm}

    \makebox[\textwidth][c]{
        \begin{subfigure}{0.48\textwidth}
            \centering
            \includegraphics[width=\textwidth]{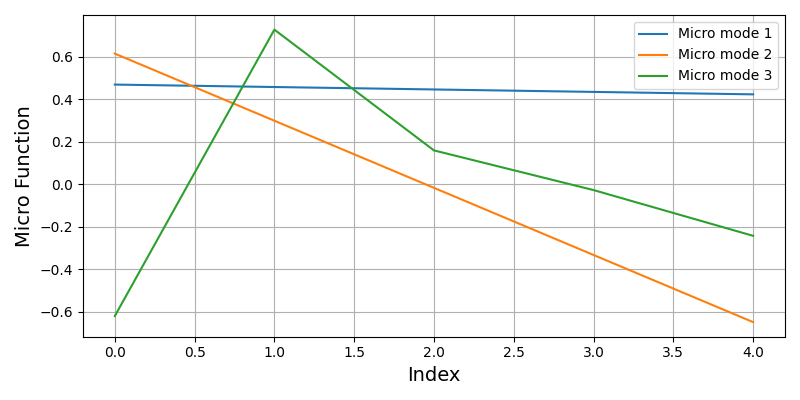}
            \caption{Micro component 1 for Eq.~\eqref{eq:duffin}.}
            \label{fig:duffin_svd3_micro1}
        \end{subfigure}
        \hspace{0.04\textwidth}
        \begin{subfigure}{0.48\textwidth}
            \centering
            \includegraphics[width=\textwidth]{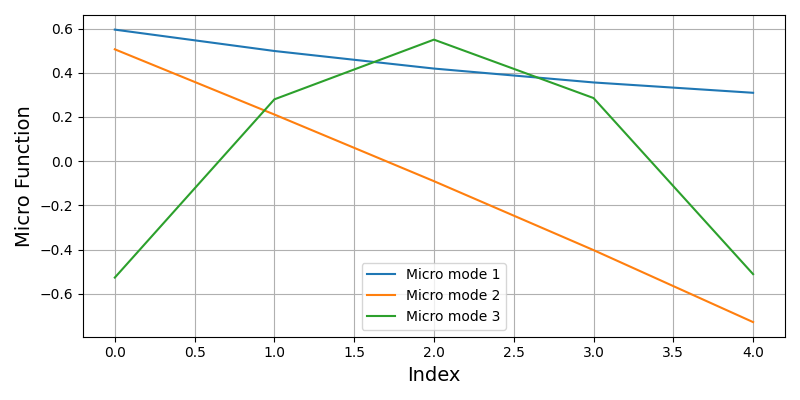}
            \caption{Micro component 2 for Eq.~\eqref{eq:duffin}.}
            \label{fig:duffin_svd3_micro2}
        \end{subfigure}
    }

    \caption{SVD approximation for Eq.~\eqref{eq:duffin} using three modes.}
    \label{fig:duffin_svd3_all}
\end{figure}

\begin{figure}[h!]
    \centering

    \begin{subfigure}{0.7\textwidth}
        \centering
        \includegraphics[width=\textwidth]{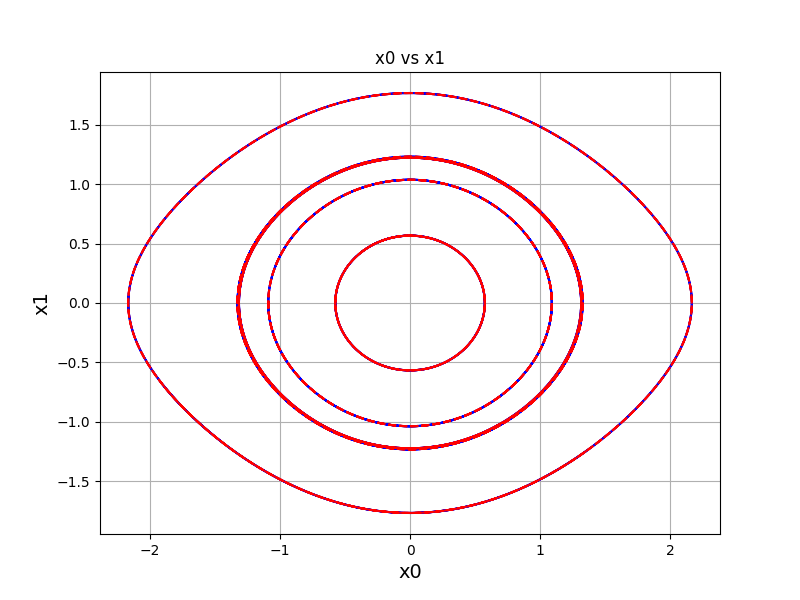}
        \caption{Function plot using three modes for Eq.~\eqref{eq:harmonic}.}
        \label{fig:harmonic_svd3_function}
    \end{subfigure}

    \vspace{0.5cm}

    \makebox[\textwidth][c]{
        \begin{subfigure}{0.48\textwidth}
            \centering
            \includegraphics[width=\textwidth]{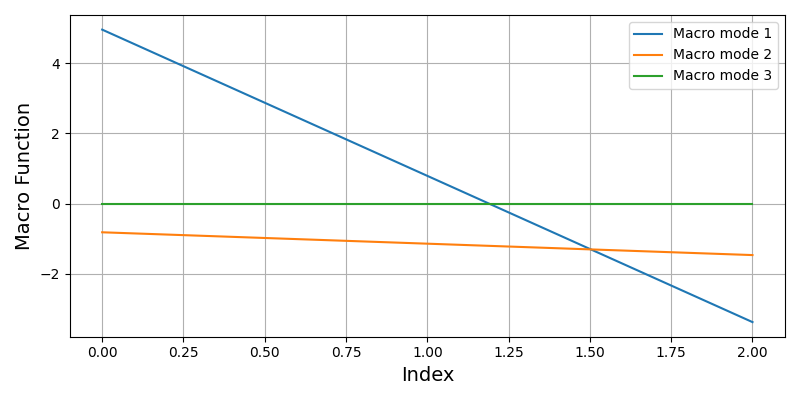}
            \caption{Macro component 1 for Eq.~\eqref{eq:harmonic}.}
            \label{fig:harmonic_svd3_macro1}
        \end{subfigure}
        \hspace{0.04\textwidth}
        \begin{subfigure}{0.48\textwidth}
            \centering
            \includegraphics[width=\textwidth]{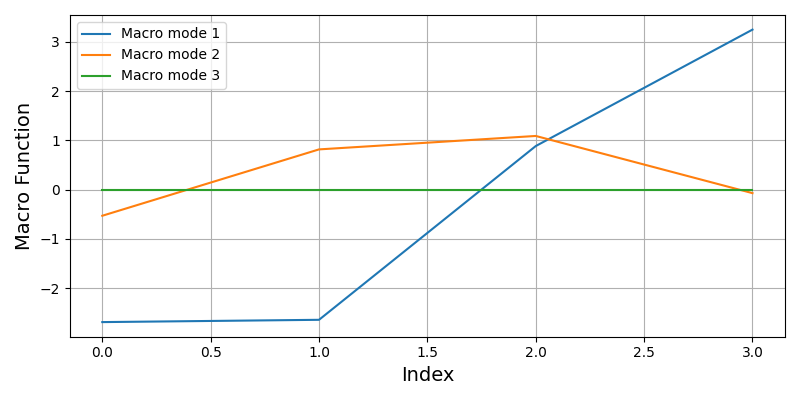}
            \caption{Macro component 2 for Eq.~\eqref{eq:harmonic}.}
            \label{fig:harmonic_svd3_macro2}
        \end{subfigure}
    }

    \vspace{0.5cm}

    \makebox[\textwidth][c]{
        \begin{subfigure}{0.48\textwidth}
            \centering
            \includegraphics[width=\textwidth]{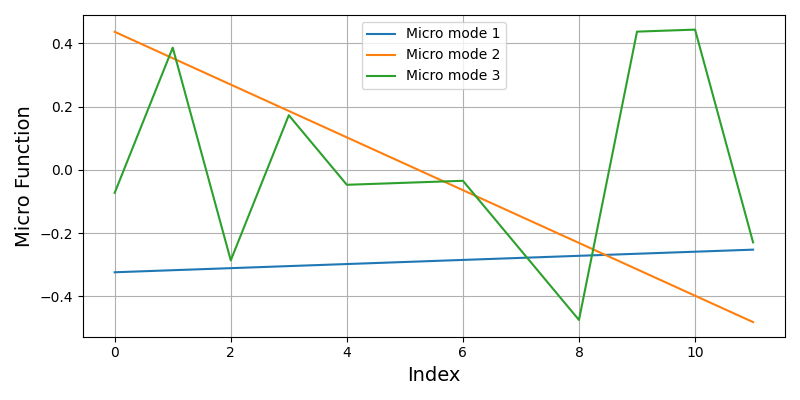}
            \caption{Micro component 1 for Eq.~\eqref{eq:harmonic}.}
            \label{fig:harmonic_svd3_micro1}
        \end{subfigure}
        \hspace{0.04\textwidth}
        \begin{subfigure}{0.48\textwidth}
            \centering
            \includegraphics[width=\textwidth]{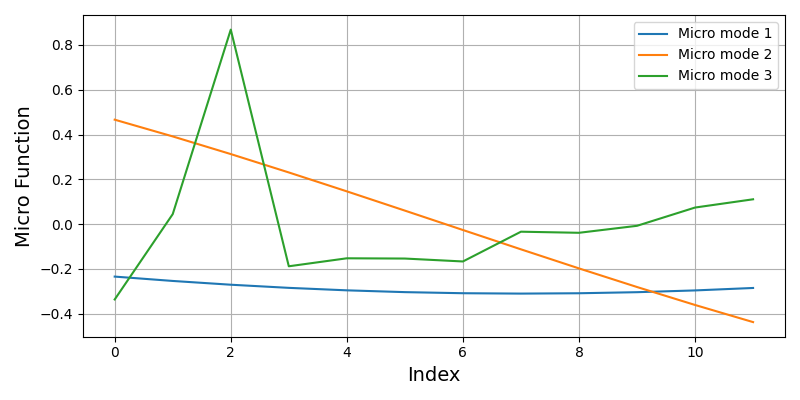}
            \caption{Micro component 2 for Eq.~\eqref{eq:harmonic}.}
            \label{fig:harmonic_svd3_micro2}
        \end{subfigure}
    }

    \caption{SVD approximation for Eq.~\eqref{eq:harmonic} using three modes.}
    \label{fig:harmonic_svd3_all}
\end{figure}

Finally, we examine the two dynamical systems given by~\eqref{eq:duffin} and~\eqref{eq:harmonic}. 
For the system described by~\eqref{eq:duffin}, the SVD approximation is shown in Figure~\ref{fig:duffin_svd3_all}. 
The function plot, together with the corresponding macro and micro components, demonstrates that the chosen number of modes captures the system dynamics with good accuracy, yielding a mean squared error (MSE) of approximately \(4.14 \times 10^{-4}\). 
Similarly, for the system described by~\eqref{eq:harmonic}, the results are presented in Figure~\ref{fig:harmonic_svd3_all}, where the SVD-based reconstruction provides an accurate representation of the dynamics with a corresponding MSE of approximately \(1.8 \times 10^{-5}\).

In summary, the numerical results demonstrate that the SVD method is a powerful approach for modeling and predicting complex, multi-scale dynamical systems. By decomposing the system into orthogonal modes and retaining only the most significant components, the method effectively reduces dimensionality while preserving essential dynamics. This enables accurate reconstruction of the system’s behavior, capturing both large-scale patterns and fine-scale variations with reduced computational cost.

\section{Sparse High-Order SVD Method}

In practical situations, it is often not feasible to access the values of a target function \( f \) at all points in the computational domain. The measurement budget may only allow a sparse sampling of the domain, leading to an incomplete spatial dataset. This sparsity poses another challenge for data-driven modeling.

A naive approach would involve approximating the function directly via a neural network:
\[
f(x) \approx NN(x),
\]
but such a general approximation involves two key limitations:
\begin{itemize}
    \item High Data Requirements: Training neural networks to approximate complex functions accurately typically requires a substantial amount of data to determine the network parameters.
    \item Risk of Overfitting: In multiscale problems, neural networks are prone to overfitting and spectral bias due to insufficient data, which impairs their ability to generalize beyond the training set, especially for high frequency data.
\end{itemize}

To address these challenges, we propose an apporach that follows a Sparse high-order SVD methodology, inspired by the PINN-PGD approach presented in \cite{math12152365}. In this manner, the algorithm is capable of learning from sparse data measurements. The scheme will follow an enrichment approach, where neural networks will learn the micro and macro functions profiting from their power to embed nonlinear correlations present in data. If the first micro and macro approximations do not accurately represent the dynamics, more  modes (more neural network approximations of the functions) will progressively be added to the representation.

\subsection{Neural SVD Representation of Sparse Observations}

Following the spirit of SVD, we begin by organizing the observed values of \( f \) into a matrix \( \boldsymbol{F} \), where each macro-cell corresponds to a single column of the matrix. However, due to the sparsity of measurements, this matrix remains incomplete, which prevents the direct application of classical SVD.

To approximate the low-rank structure of \( \boldsymbol{F} \) despite its sparsity, we model its decomposition using time multiscale dimensional decomposition \cite{math12152365}:
\[
f(x_{i},x_{j}) \approx NN_U(x_i) \cdot NN_V(x_j) 
\]
We further define:
\begin{itemize}
    \item \( U_i = NN_U(x_i) \), where \( x_i \) denotes the spatial coordinates within the macro-cell (i.e., the row index of the matrix),
    \item \( V_j = NN_V(x_j) \), where \( x_j \) denotes the macro-cell identifier or coordinates (i.e., the column index).
\end{itemize}

The training loss is defined as \cite{math12152365}:
\[
\mathcal{L}_{\text{data}} = \sum_{(i,j) \in \Omega} \left( f(x_{i},x_{j})) - NN_U(x_i) \cdot NN_V(x_j)\right)^2,
\]
where \( \Omega \subset \{(i,j)\} \) denotes the set of sparse observed data points.

\subsection{Residual Correction}

To enhance this data-driven model with physical consistency, we adopt a residual-based enrichment strategy. After the first approximation is learned, we define a residual:
\[
r^{(1)}(x_{i},x_{j}) = f(x_{i},x_{j}) -  NN_U^{(1)}(x_i) \cdot NN_V^{(1)}(x_j),
\]
and fit a second neural network pair \( (NN_U^{(2)}, NN_V^{(2)}) \) to approximate this residual:
\[
\mathcal{L}^{(2)} = \sum_{(i,j) \in \Omega} \left( r^{(1)}(x_{i},x_{j}) -  NN_U^{(2)}(x_i)\cdot NN_V^{(2)}(x_j)  \right)^2,
\]

This process can be repeated iteratively, resulting in a parsimonious enrichment:
\[
\hat{f}(x_{i},x_{j}) = \sum_{k=1}^{K} \langle NN_U^{(k)}(x_i), NN_V^{(k)}(x_j) \rangle.
\]

This approximation allows us to flexibly capture multiscale dynamics in a data-efficient and physics-consistent manner by decomposing the solution into a sum of learned low-rank components.

\subsection{Numerical Results}

To evaluate the Sparse High-Order SVD framework, we apply it to the same representative one-dimensional function exhibiting multiscale characteristics defined in Equation~\eqref{eq:cos&sin}. We randomly mask 70\% of the function values across the domain, producing a highly sparse observation matrix \( \boldsymbol{F} \), with only 30\% of its entries known. The goal is to reconstruct the full matrix, including the missing entries, using the Sparse High-Order SVD framework.


Following the formulation, we employ two separate neural networks to learn a low-rank approximation of the matrix:
\begin{itemize}
    \item \( NN_U(x_i) \) predicts the latent representation of the micro-scale component (rows of \( \boldsymbol{F} \)),
    \item \( NN_V(x_j) \) predicts the latent representation of the macro-scale component (columns of \( \boldsymbol{F} \)).
\end{itemize}

Each neural network consists of three hidden layers with nonlinear activation functions. The model is trained to minimize the reconstruction loss over the training observed entries.

\begin{figure}[h!]
    \centering
    \begin{subfigure}{0.7\textwidth}
        \centering
        \includegraphics[width=\linewidth]{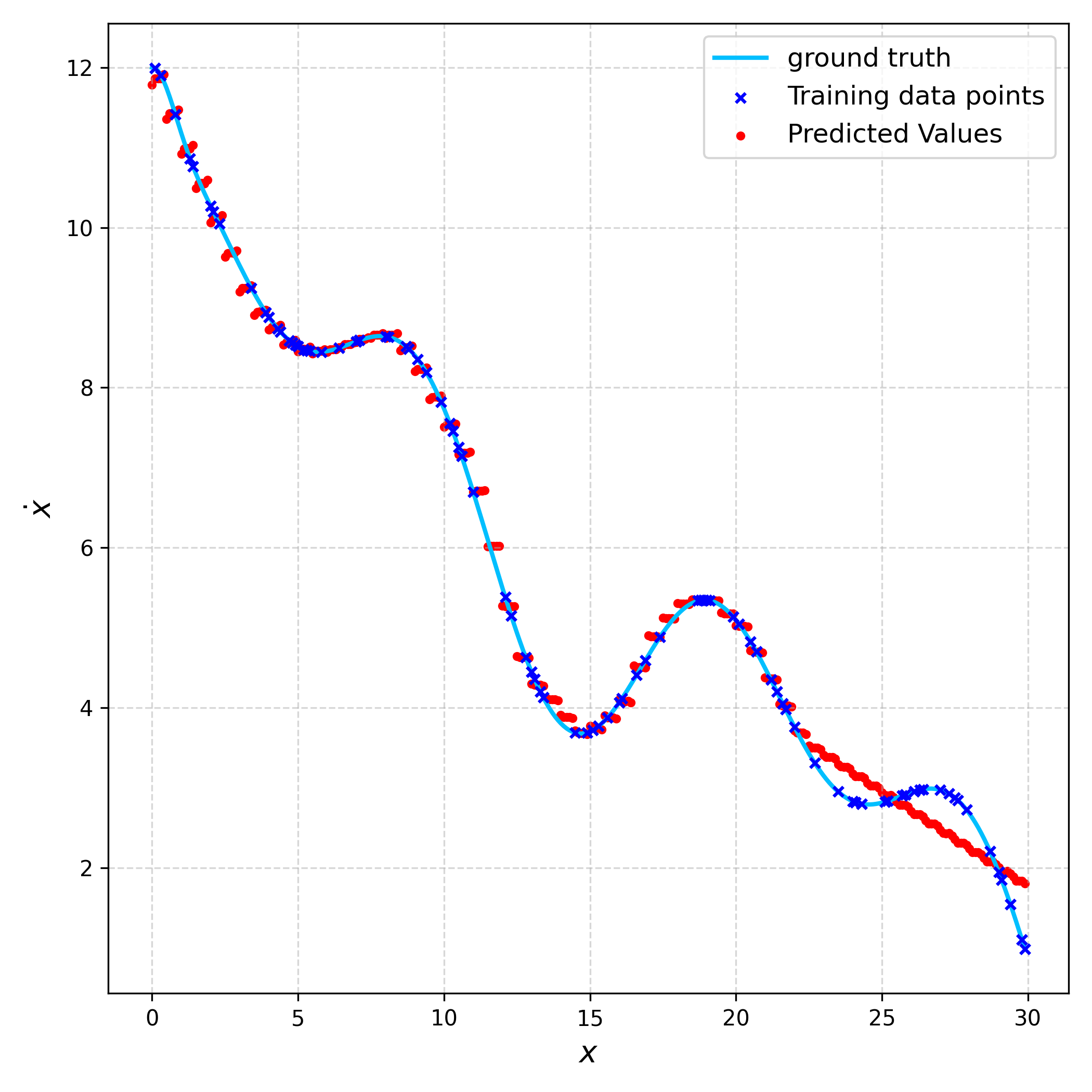}
        \caption{Reconstructed function.}
        \label{fig:stage1_full}
    \end{subfigure}

    \vspace{0.8em} 

    \begin{subfigure}{0.48\textwidth}
        \centering
        \includegraphics[width=\linewidth]{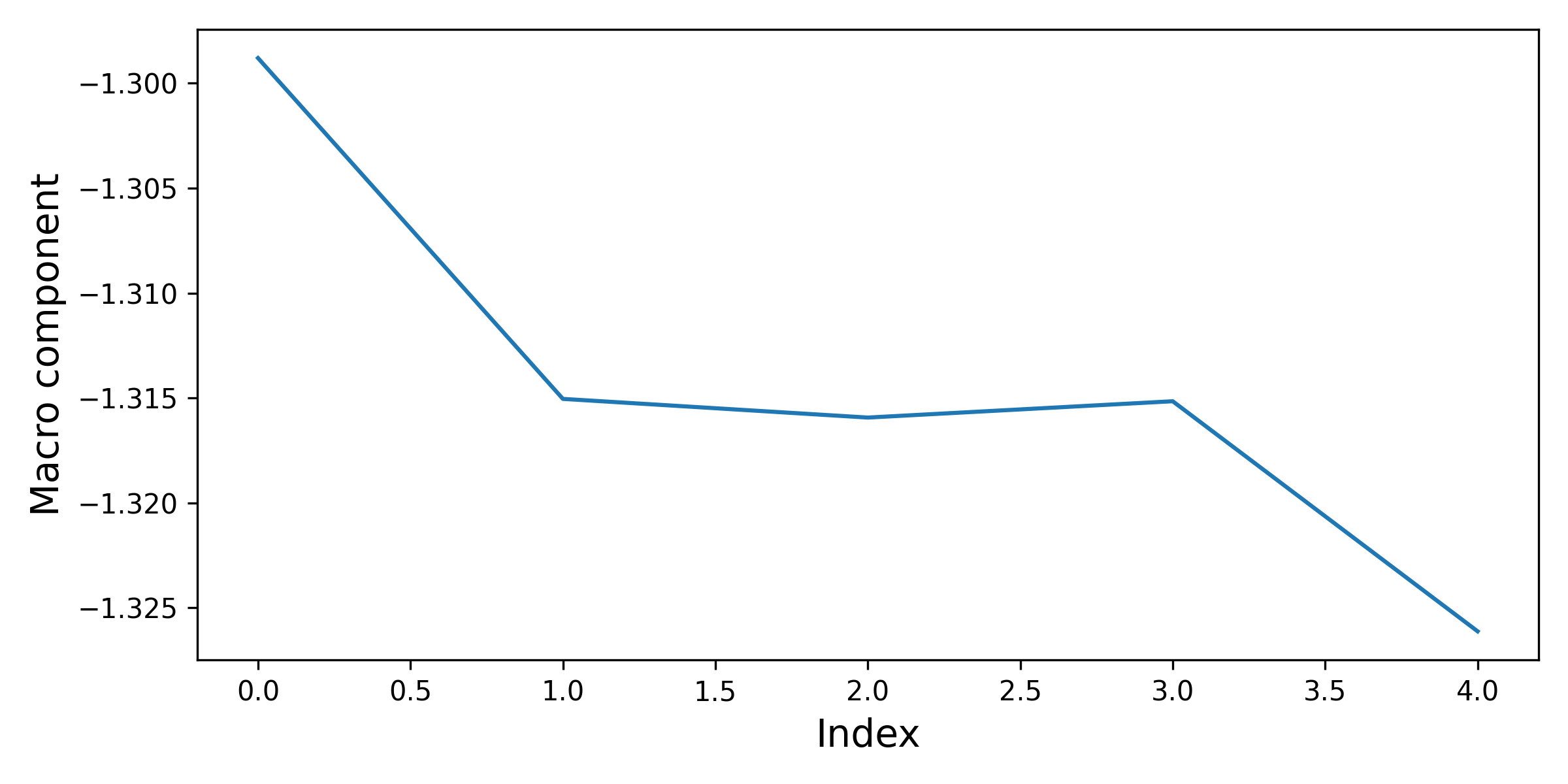}
        \caption{Macro component.}
        \label{fig:stage1_macro}
    \end{subfigure}
    \hfill
    \begin{subfigure}{0.48\textwidth}
        \centering
        \includegraphics[width=\linewidth]{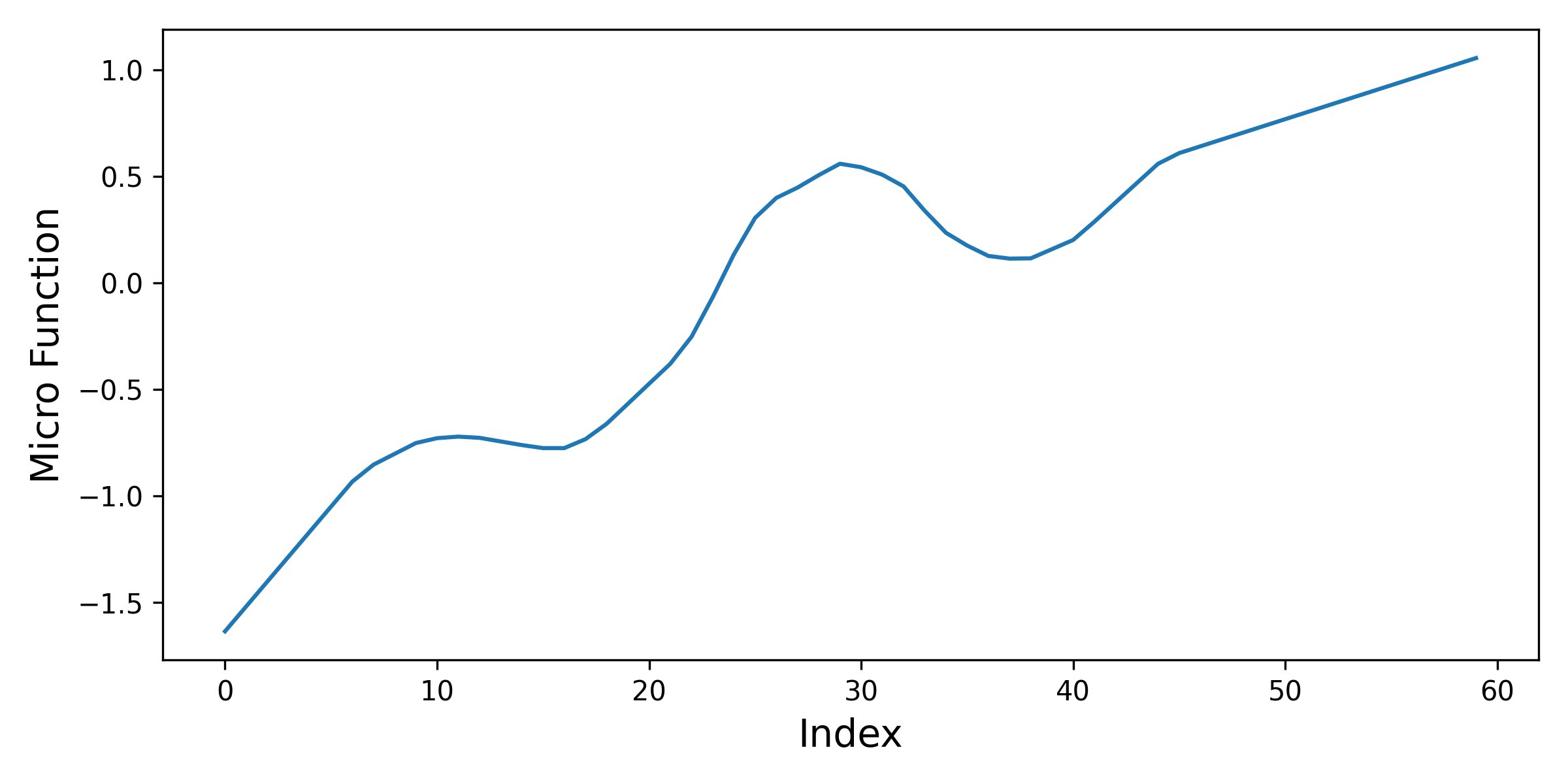}
        \caption{Micro component.}
        \label{fig:stage1_micro}
    \end{subfigure}

    \caption{Sparse High-Order SVD with a single macro–micro decomposition (\( \text{Stage} = 1 \)).}
    \label{fig:stage1_reconstruction}
\end{figure}

\begin{figure}[h!]
    \centering
    \begin{subfigure}{0.7\textwidth}
        \centering
        \includegraphics[width=\linewidth]{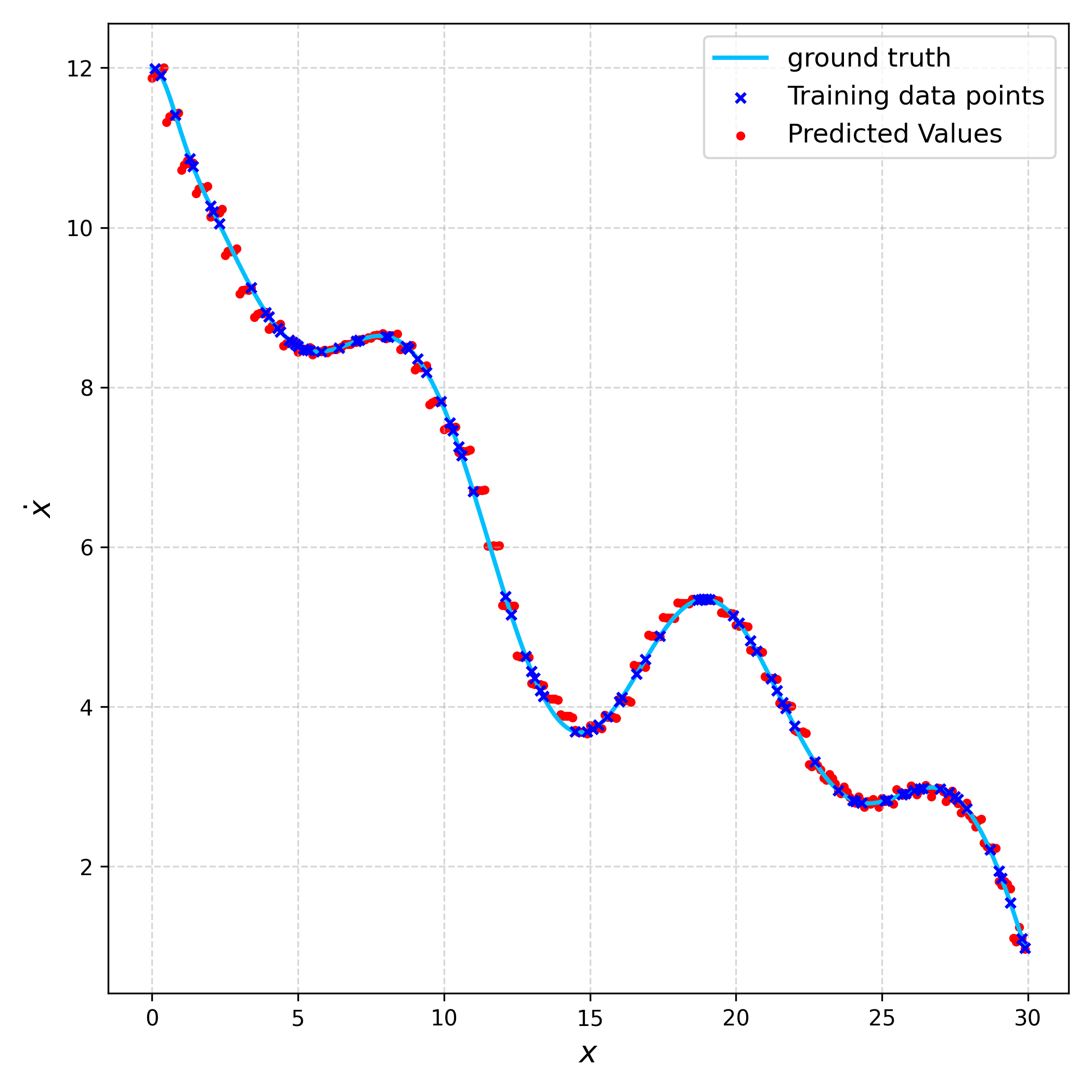}
        \caption{Reconstructed function.}
        \label{fig:stage2_full}
    \end{subfigure}

    \vspace{0.8em} 

    \begin{subfigure}{0.48\textwidth}
        \centering
        \includegraphics[width=\linewidth]{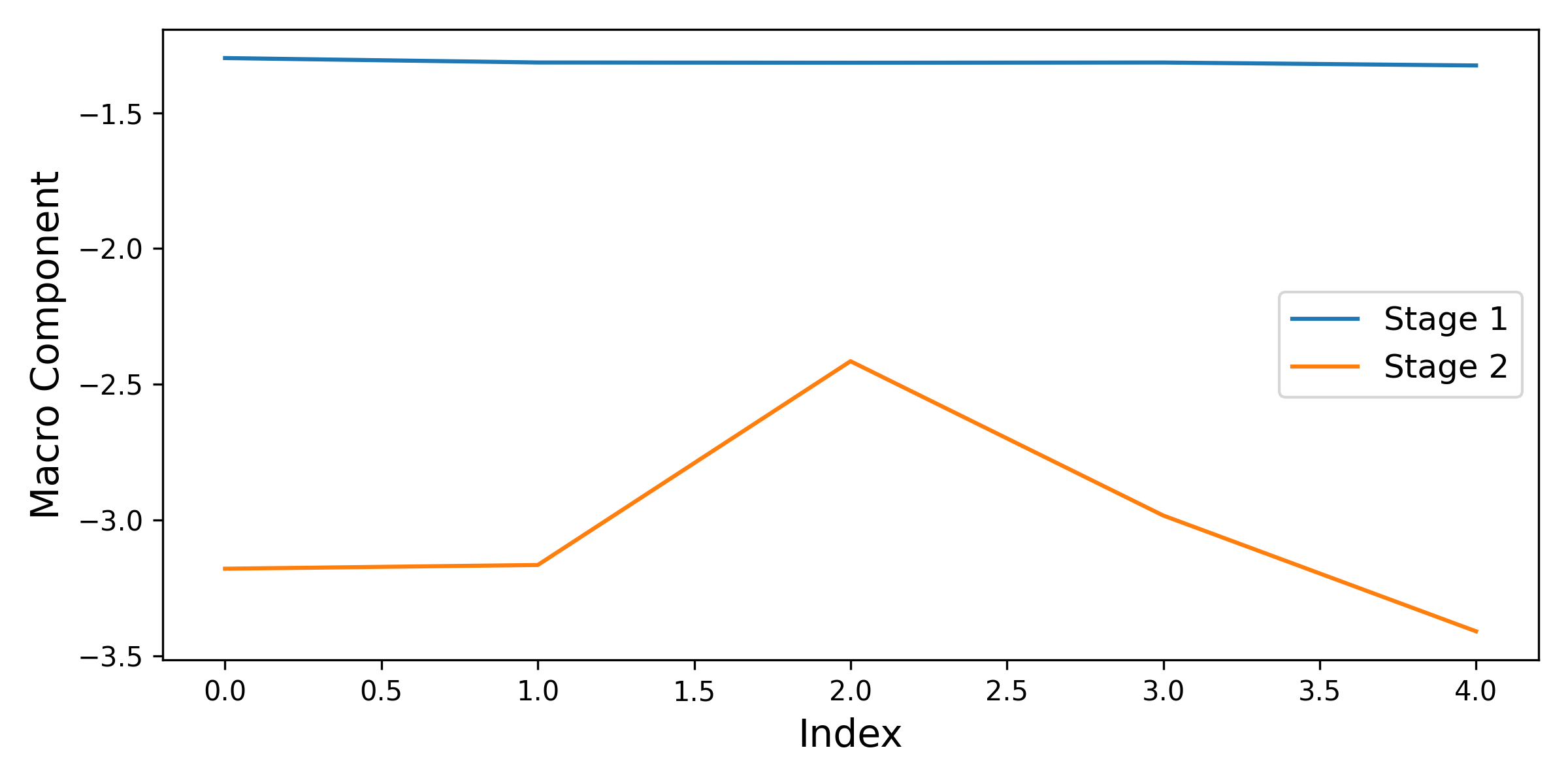}
        \caption{Macro component.}
        \label{fig:stage2_macro}
    \end{subfigure}
    \hfill
    \begin{subfigure}{0.48\textwidth}
        \centering
        \includegraphics[width=\linewidth]{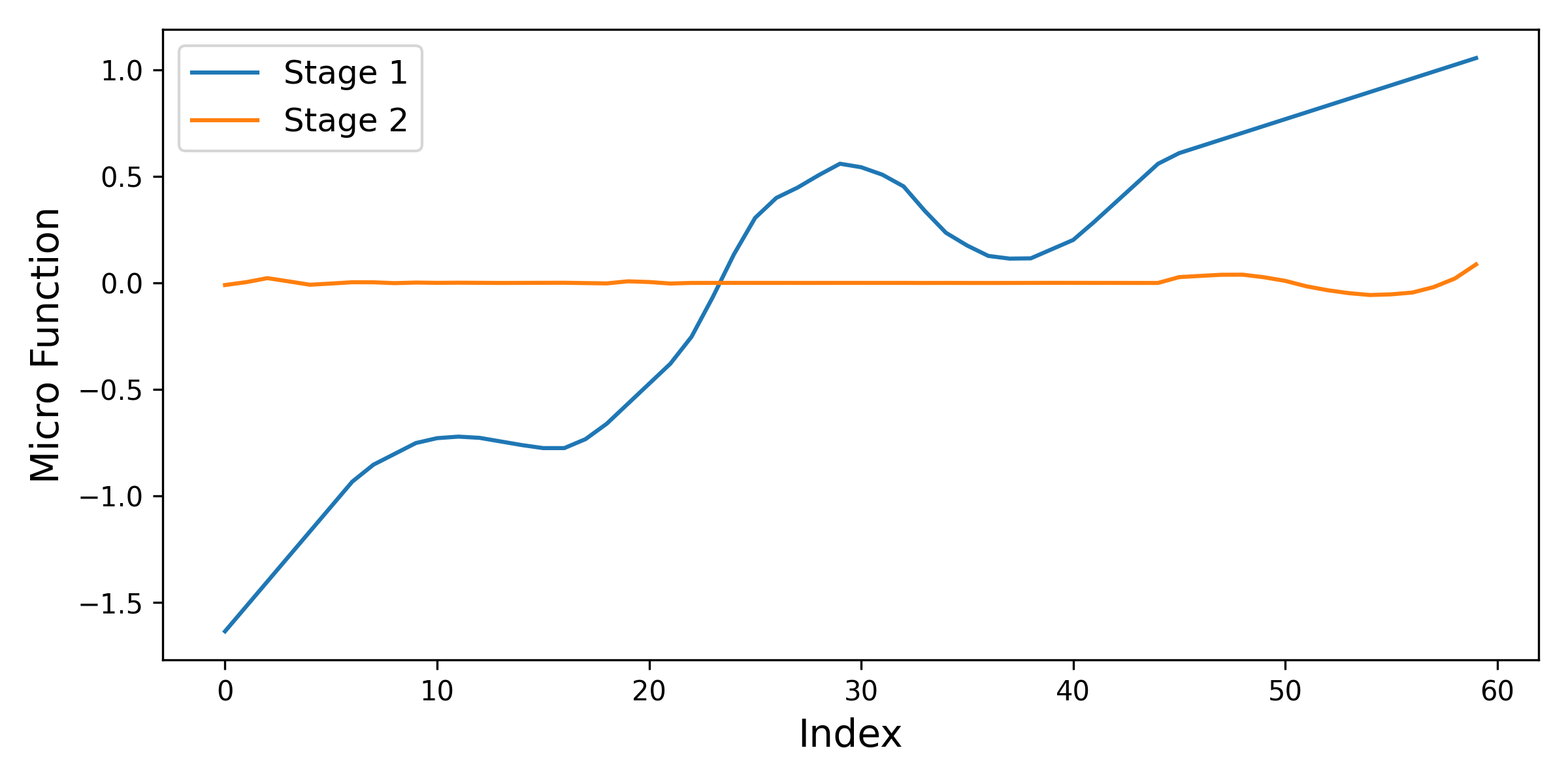}
        \caption{Micro component.}
        \label{fig:stage2_micro}
    \end{subfigure}

    \caption{Sparse High-Order SVD with residual enrichment (\( \text{Stage} = 2 \)).}
    \label{fig:stage2_reconstruction}
\end{figure}

In this example, we tested the Sparse High-Order SVD model with two configurations: Stage~1 (single macro--micro decomposition) and Stage~2 (residual enrichment with an additional decomposition). In the Stage~1 setting, as shown in Figure~\ref{fig:stage1_reconstruction}, the model already provided a good-quality reconstruction, demonstrating that the macro and micro components effectively capture the coarse- and fine-scale dynamics of the signal, with a mean squared error of 0.08.  

However, as shown in Figure~\ref{fig:stage2_reconstruction}, the Stage~2 configuration with residual enrichment achieved clearly more accurate results. The additional stage improved the agreement with the true signal, refining both the macro-scale trends and the micro-scale oscillations, and reducing the MSE to 0.0007.  

This numerical example confirms the ability of the Sparse High-Order SVD method to perform low-rank reconstruction of functions from sparse measurements in time.

\section{Conclusion}

In this work, we proposed several approaches for analyzing and predicting the behavior of dynamical systems by decomposing their dynamics into micro (fine-scale) and macro (coarse-scale) components. One approach leverages the Partition of Unity (PU) method combined with neural networks to construct localized approximations, enabling the prediction of both macro- and micro-scale behaviors. Another approach uses Singular Value Decomposition (SVD) to extract dominant modes from the data, providing smooth global basis functions that capture the principal features of the system’s dynamics across scales. To address scenarios with sparse or incomplete measurements, we further employ a Sparse High-Order SVD, which enables the reconstruction of multiscale dynamics directly from limited observations.

These approaches were applied to a variety of representative dynamical systems, including polynomial, sinusoidal, and exponential models, demonstrating their generality and robustness. The results confirmed that these methods effectively isolate and capture multi-scale features, leading to improvements in prediction accuracy and computational efficiency. Incorporating micro-macro decomposition with data-driven techniques provides a systematic way to model complex dynamics while maintaining interpretability and scalability, which is essential for high-dimensional and real-time applications.

Beyond predictive performance, these approaches contribute to a deeper understanding of how multi-scale phenomena can be disentangled and represented systematically. They also open avenues for integrating domain knowledge with data-driven techniques, enabling hybrid modeling approaches that balance physical interpretability and flexibility.

Future work will focus on extending these methods to more complex and realistic systems—such as turbulent flows, biological networks, or large-scale infrastructure systems—and improving performance by integrating with advanced machine learning architectures, including neural ODEs, physics-informed neural networks, or transformer-based sequence models. Additionally, adaptive schemes that dynamically adjust the level of resolution or partitioning based on data characteristics could further enhance model performance and generalizability. The integration of PU and SVD into broader modeling pipelines offers a promising direction for building intelligent, adaptive, and efficient models capable of addressing the challenges posed by real-world dynamical systems.

\section*{Acknowledgements}

This work was supported by the IN-DEEP European Project. This Project has received funding from the EU's Horizon Europe research and innovation programme under the Marie Sklodowska-Curie GA No 101119556.

This research is also part of the DesCartes programme and is supported by the National Research Foundation, Prime Minister Office, Singapore under its Campus for Research Excellence and Technological Enterprise (CREATE) programme.

We acknowledge support from the French government, managed by the National Research Agency (ANR), under the CPJ ITTI.

We would like to thank the research teams at PIMM Laboratory for their valuable insights and technical support throughout this project.

\bibliographystyle{elsarticle-num}

\end{document}